\documentclass[10pt,twocolumn,letterpaper]{article}
\usepackage[pagenumbers]{cvpr}
\usepackage{amsmath,amssymb}
\usepackage{booktabs}
\usepackage[dvipsnames]{xcolor}
\usepackage[hyphens]{url}
\usepackage{graphicx}
\usepackage{wrapfig}

\AtBeginDocument{%
  }

\usepackage{tabularray}
\usepackage{placeins}
\usepackage{float}
\usepackage{comment}
\usepackage{microtype}

\definecolor{cvprblue}{rgb}{0.21,0.49,0.74}
\usepackage[breaklinks,colorlinks,allcolors=cvprblue]{hyperref}
\hypersetup{
  hypertexnames=false,
  pdftitle={3DTexMOR: 3D Gaussian Multi-Object Removal via Texture-Space Inpainting},
  pdfauthor={Kunxin Guang; Yonghao Zhao; Beibei Wang},
  pdfsubject={3D scene editing and multi-object removal}
}

\excludecomment{CCSXML}
\providecommand{\ccsdesc}[2][]{ }
\providecommand{\keywords}[1]{ }
\providecommand{\Description}[1]{ }

\begin{document}

\title{3DTexMOR: 3D Gaussian Multi-Object Removal via Texture-Space Inpainting}

\author{%
Kunxin Guang\textsuperscript{1,*}\quad
Yonghao Zhao\textsuperscript{2,*}\quad
Beibei Wang\textsuperscript{1,\ensuremath{\dagger}}\\[0.4em]
\normalfont\textsuperscript{1}Nanjing University\qquad
\normalfont\textsuperscript{2}Nankai University\\[0.3em]
\normalfont\small\textsuperscript{*}Equal contribution.\quad
\textsuperscript{\ensuremath{\dagger}}Corresponding author.
}

\twocolumn[{%
\maketitle
\begin{minipage}{\textwidth}
  \centering
  \captionsetup{hypcap=false}
  \includegraphics[width=\linewidth]{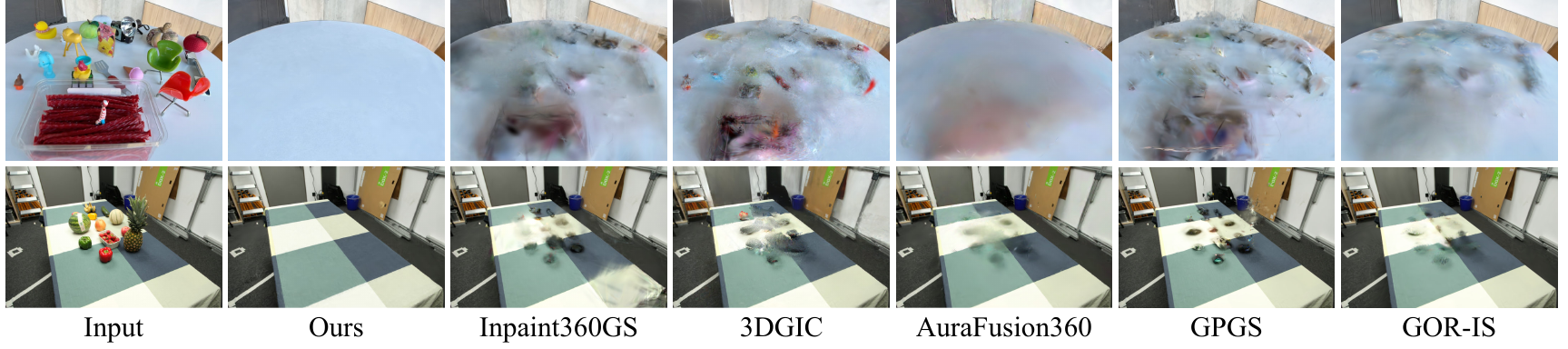}
  \captionof{figure}{
  We propose 3DTexMOR, a novel framework for high-quality 3D multi-object removal. By performing inpainting in a unified texture space, 3DTexMOR recovers the regions occluded by the removed objects with consistent appearance and geometry. Compared with existing methods, 3DTexMOR produces more consistent completion scenes with fewer artifacts and less blurriness.
  }
  \Description{A teaser figure showing multi-object removal examples for 3DTexMOR
  on tabletop scenes, with cluttered inputs and completed object-free outputs.
  }
  \label{fig:teaser}
\end{minipage}
\vspace{1em}
}]

\begin{abstract}
3D object removal aims to remove target objects from reconstructed scenes while completing the geometry and appearance of the regions they occlude. Existing methods built on Neural Radiance Fields (NeRF) or 3D Gaussian Splatting (3DGS) typically inpaint 2D images and use the completed views to guide 3D scene completion. While these methods achieve promising results in scenes with few target objects, they often struggle with complex multi-object layouts, where individual views capture only part of the surrounding texture and structure needed to guide inpainting. These limited visual cues make 2D inpainting prone to artifacts. Moreover, appearance inconsistencies across inpainted views can introduce conflicting supervision for 3D scene completion, leading to blurry reconstructions.
To address these challenges, we propose 3D Gaussian Multi-Object Removal via Texture-Space Inpainting (\textbf{3DTexMOR}). Our key idea is to perform inpainting in a unified texture space that combines complementary observations across views, providing richer visual information for recovering missing regions while promoting cross-view appearance consistency. Specifically, we introduce a texture-space inpainting module that aggregates multi-view observations into texture maps and reprojects the completed maps into camera views to supervise 3D Gaussian scene completion. Since direct RGB aggregation is unreliable on glossy surfaces, we further decompose appearance and aggregate view-independent intrinsic attributes in texture space. Finally, because completed textures alone do not sufficiently constrain 3D structure, we introduce a geometrically regularized Gaussian completion module to encourage plausible geometry in the completed regions.
Extensive experiments demonstrate that 3DTexMOR produces visually plausible scene completions and achieves state-of-the-art performance on multi-object removal. On the Syn-Diffuse dataset, it improves full-image PSNR by \textbf{6.27~dB} and reduces LPIPS by \textbf{58.6\%} relative to the strongest baseline.
\end{abstract}

\ccsdesc[500]{Computing methodologies~Computer graphics}
\keywords{3D Gaussian Splatting, object removal, scene inpainting}

\section{Introduction}
\label{sec:intro}
                                                                        Given multi-view images, 3D object removal aims to reconstruct a 3D scene, remove specified objects, and complete the regions they occlude with plausible geometry and appearance. This capability supports flexible scene editing for applications such as embodied intelligence and virtual reality. Recent advances in 3D scene representations, including Neural Radiance Fields~(NeRF)~\citep{mildenhall2020nerf} and 3D Gaussian Splatting~(3DGS)~\citep{kerbl2023gaussian}, have provided a foundation for reconstructing and editing such scenes. However, recovering the occluded regions remains challenging, as the completed content must be visually plausible and consistent across viewpoints.

Existing methods for 3D object removal typically rely on image-space inpainting, with different strategies for constructing multi-view supervision. One strategy directly inpaints multiple camera views and uses the completed images to optimize the 3D scene~\citep{weder2023removing,mirzaei2023spinnerf,weber2024nerfiller,chen2024gaussianeditor,wang2024gscream,seo2025repaintgs}. This provides completion supervision from multiple viewpoints, but the inpainted images may contain inconsistent appearances for the same scene regions. Another strategy completes a few reference views, in either RGB or intrinsic space, and propagates the results to other views through depth-based reprojection~\citep{huang2025_3dgic,zhao2026goris}. Sharing completed content across views helps reduce appearance inconsistencies, but the accuracy of the resulting supervision depends on the geometry used for reprojection, and errors in the completed regions can lead to incorrect supervision. Recent methods further exploit diffusion priors to improve the visual quality of image completion~\citep{wu2025aurafusion360,wang2026inpaint360gs}, while retaining their reliance on image-space inpainting. Despite these advances, two challenges remain in complex multi-object scenes. First, individual views often capture only part of the surrounding texture and structure needed to guide inpainting, leaving limited visual information for recovering the missing regions and making completion prone to artifacts. Second, maintaining appearance consistency across inpainted views remains difficult, and the resulting conflicting supervision can lead to blurry 3D reconstructions.

To address these challenges, we propose 3D Gaussian Multi-Object Removal via Texture-Space Inpainting (\textbf{3DTexMOR}). Our key idea is to perform inpainting in a unified texture space shared across camera views. By combining complementary observations from multiple views, this space brings together the visible texture and structure surrounding the missing regions, providing richer visual information for inpainting. Moreover, completing the missing regions in this shared space allows different views to receive supervision from the same completed texture maps, reducing appearance inconsistencies during 3D scene completion. Specifically, we propose a texture-space inpainting module that aggregates multi-view observations into texture maps for inpainting and reprojects the completed maps into camera views to provide shared supervision for 3D scene completion. To reduce aggregation errors caused by view-dependent appearance on glossy surfaces, the module operates on view-independent intrinsic attributes obtained through the appearance decomposition of GOR-IS~\citep{zhao2026goris}. We also propose a geometrically regularized Gaussian completion module to address the insufficient geometric constraints provided by completed textures alone. Experiments on synthetic and real-world scenes demonstrate that 3DTexMOR produces visually plausible scene completions under both single- and multi-object removal settings. On synthetic diffuse multi-object scenes, 3DTexMOR improves full-image PSNR by 6.27~dB and reduces LPIPS by 58.6\% compared with the strongest baseline. Our main contributions are as follows:
\begin{itemize}
    \item \textbf{3DTexMOR}, a framework for 3D Gaussian multi-object removal that combines complementary observations across views through unified texture-space inpainting,
    \item a texture-space inpainting module that aggregates multi-view intrinsic images into texture maps for inpainting, providing shared supervision across views, and
    \item a geometrically regularized Gaussian completion module that complements texture supervision with geometric constraints to recover plausible 3D structure.
\end{itemize}

\section{Related work}
\label{sec:relatedwork}

\paragraph{Object-level understanding and editing in 3D scenes.}
Understanding the object composition of a scene helps reliably separate individual objects. Promptable 2D segmentation and open-vocabulary perception models, such as SAM~\citep{kirillov2023segment}, have therefore been extended to 3D through feature distillation and multi-view mask propagation. LERF~\citep{kerr2023lerf} embeds language features into NeRF for open-ended 3D queries, while SA3D~\citep{cen2023sa3d} propagates SAM prompts through NeRF for 3D object segmentation. Feature 3DGS~\citep{zhou2024feature3dgs} and LangSplat~\citep{qin2024langsplat} further distill high-dimensional semantic or language features into 3D Gaussians for efficient segmentation and querying. More closely related to editable Gaussian scenes, SAGA~\citep{cen2025saga} and Gaussian Grouping~\citep{ye2024gaussiangrouping} attach segmentation or object identity features to Gaussians, enabling object grouping, selection, and local editing. 

\paragraph{3D inpainting and object removal.}
The key challenge of 3D object removal is to recover occluded regions with plausible appearance and geometry. Existing 2D inpainting methods~\citep{suvorov2022lama,lugmayr2022repaint} provide strong image priors. However, when applied independently to different views, they may produce inconsistent appearances for the same regions. Building on NeRF and 3DGS, several methods extend 2D inpainting to 3D scenes. NeRF-based object removal methods~\citep{weder2023removing,mirzaei2023spinnerf,weber2024nerfiller} combine multi-view masks, 2D inpainting supervision, and radiance-field optimization. 3DGS-based methods further improve the efficiency and quality of 3D scene inpainting. GaussianEditor~\citep{chen2024gaussianeditor} enables controllable Gaussian editing with diffusion guidance, while GScream~\citep{wang2024gscream} improves geometry and feature consistency for Gaussian object removal. Recent works further explore reference-guided Gaussian scene completion~\citep{seo2025repaintgs,wu2025aurafusion360}, depth-guided Gaussian scene inpainting~\citep{huang2025_3dgic,dahaghin2025splatfill}, geometry-aware completion with projected image refinement~\citep{lee2026gpgs}, 360-degree multi-object Gaussian inpainting~\citep{wang2026inpaint360gs}, shadow-aware completion~\citep{zheng2026gaucomp}, and residual artifact removal~\citep{kocour2025remove360}. InstaInpaint~\citep{you2025instainpaint} enables efficient feed-forward scene inpainting through masked reconstruction conditioned on posed images, multi-view masks, and a reference proposal. GOR-IS~\citep{zhao2026goris} improves physical consistency by decomposing scenes into intrinsic components, modeling light transport, and performing inpainting in intrinsic space. Nevertheless, most existing methods still rely on image-based inpainting and mainly focus on single-object removal, making it difficult to handle complex multi-object scenes while maintaining consistent geometry and appearance in the completed 3D scenes.

\section{Method}
\label{sec:method}

\subsection{Overview of the 3DTexMOR Framework}
\label{subsec:problem-overview}

\begin{figure*}[t]
  \centering
  \includegraphics[width=\textwidth]{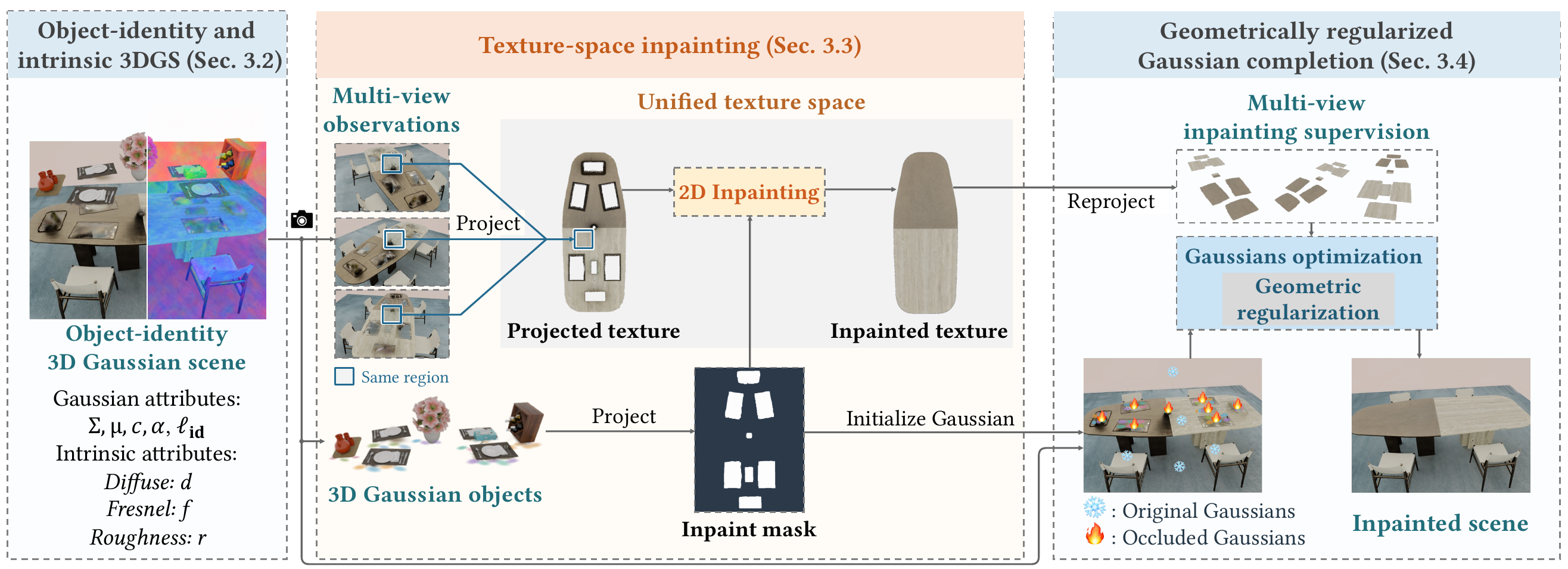}
  \caption{\textbf{Overview of the 3DTexMOR framework.} Our Gaussian scene representation incorporates object identity features and intrinsic appearance attributes. The texture-space inpainting module aggregates complementary multi-view intrinsic observations into a shared texture space, completes the masked regions, and reprojects the completed maps to provide shared supervision across camera views. The geometrically regularized Gaussian completion module lifts the completed textures into a 3D Gaussian representation on the estimated support plane and regularizes its geometry during optimization. The final scene combines the original Gaussians representing the unremoved scene content with the optimized completion Gaussians.}
  \label{fig:pipeline}
\end{figure*}

3DTexMOR removes specified objects from a reconstructed 3D Gaussian scene and completes the occluded regions with plausible geometry and appearance. As illustrated in Fig.~\ref{fig:pipeline}, our framework builds on a Gaussian representation augmented with object identity features for selective removal and view-independent intrinsic attributes for multi-view aggregation (Sec.~\ref{subsec:object-intrinsic-decomposition}). Based on this representation, we introduce two modules. The \textbf{texture-space inpainting module} aggregates complementary multi-view intrinsic observations into a unified texture space, inpaints the missing regions, and reprojects the completed maps to the camera views as shared supervision for 3D scene completion (Sec.~\ref{subsec:reversible-texture-mapping}). The \textbf{geometrically regularized Gaussian completion module} completes the regions previously occluded by the removed objects, using the reprojected intrinsic supervision to recover their appearance and geometric constraints to encourage plausible geometry (Sec.~\ref{subsec:scene-completion-optimization}).

\subsection{Object-identity and intrinsic 3DGS}
\label{subsec:object-intrinsic-decomposition}

Our scene representation augments 3D Gaussians with object identity features and intrinsic appearance attributes. Object identity features support selective removal, while intrinsic attributes provide view-independent appearance information for aggregation across views.

\textbf{Object-identity representation.}
Given multi-view images, we reconstruct a scene represented by a set of 3D Gaussians $\mathcal{G}$ using 3DGS~\citep{kerbl2023gaussian}. We augment each Gaussian $g_i$ with a learnable object identity feature $\ell_\mathbf{id}$. These features are rendered into the training views and supervised using multi-object masks to encourage consistent object identities across views. Let $\mathcal{O}_{\mathrm{rm}}$ denote the set of object labels selected for removal. We remove the Gaussians associated with these labels and retain the remaining set: $\mathcal{G}_{\mathrm{keep}}=\left\{g_i\in\mathcal{G}\mid\ell_\mathbf{id}\notin\mathcal{O}_{\mathrm{rm}}\right\}.$ Here, $\mathcal{G}_{\mathrm{keep}}$ contains the original Gaussians representing the background and objects not selected for removal. The regions previously occluded by the removed objects require subsequent geometry and appearance completion.

\textbf{Intrinsic decomposition.}
Directly aggregating multi-view RGB observations is unreliable on glossy surfaces because their appearance varies with viewing direction. Following GOR-IS~\citep{zhao2026goris}, we decompose scene appearance into a view-independent diffuse component $D$ and a view-dependent glossy component $G$: $C(\mathbf{x},\boldsymbol{\omega}_o)=D(\mathbf{x}) + G(\mathbf{x},\boldsymbol{\omega}_o).$ Here, $\mathbf{x}$ denotes a surface position and $\boldsymbol{\omega}_o$ denotes the outgoing viewing direction. Each Gaussian $g_i$ carries diffuse radiance $\mathbf{d}_i$, Fresnel reflectance $\mathbf{f}_i$, and roughness $r_i$. These attributes are splatted into each camera view $c$ to obtain the intrinsic maps $X_c=\{D_c, F_c, R_c\}$. The diffuse map $D_c$ provides the diffuse component, while the glossy component is computed through physically-based rendering using $F_c$ and $R_c$, together with scene geometry and lighting. Further rendering details are provided in Appendix~\ref{app:reflection-details} of the supplementary material. The resulting intrinsic maps provide observations of the underlying view-independent attributes for subsequent aggregation and inpainting in the unified texture space (Sec.~\ref{subsec:reversible-texture-mapping}).

\subsection{Texture-Space Inpainting}
\label{subsec:reversible-texture-mapping}
Individual camera views often reveal only part of the appearance information surrounding the missing regions. We therefore aggregate complementary multi-view intrinsic observations into a unified texture space before inpainting. Completing the missing regions in this shared space and reprojecting the results to the camera views provides supervision derived from the same completed textures, reducing cross-view appearance inconsistencies.

\textbf{Texture space construction.}
We construct the texture space on an estimated support plane, assuming that the regions to be completed lie approximately on a common plane. We estimate this plane from background Gaussian centers using least-squares fitting and represent it as $\mathbf{n}^{\top}\mathbf{x}+d=0$, where $\mathbf{n}$ is the unit normal and $d$ is the plane offset.

To parameterize the plane, we select a reference point $\mathbf{o}$ on it as the origin and a unit tangent vector $\mathbf{e}_u$ orthogonal to $\mathbf{n}$ and define the second tangent axis as $\mathbf{e}_v=\mathbf{n}\times\mathbf{e}_u$. The mapping from 3D positions to texture coordinates is: $T(\mathbf{x})=\big[(\mathbf{x}-\mathbf{o})^{\top}\mathbf{e}_u,\,(\mathbf{x}-\mathbf{o})^{\top}\mathbf{e}_v\big]^{\top}.$
This mapping assigns 3D locations to coordinates in the unified texture space. For each pixel $p$ in camera view $c$, we intersect its viewing ray with the estimated support plane to obtain the 3D point $\mathbf{x}_c(p)$ and texture coordinate $\mathbf{u}_c(p)=T(\mathbf{x}_c(p))$. Observations occluded by retained scene content are excluded.

\begin{figure}[t]
    \centering
    \includegraphics[width=\linewidth]
        {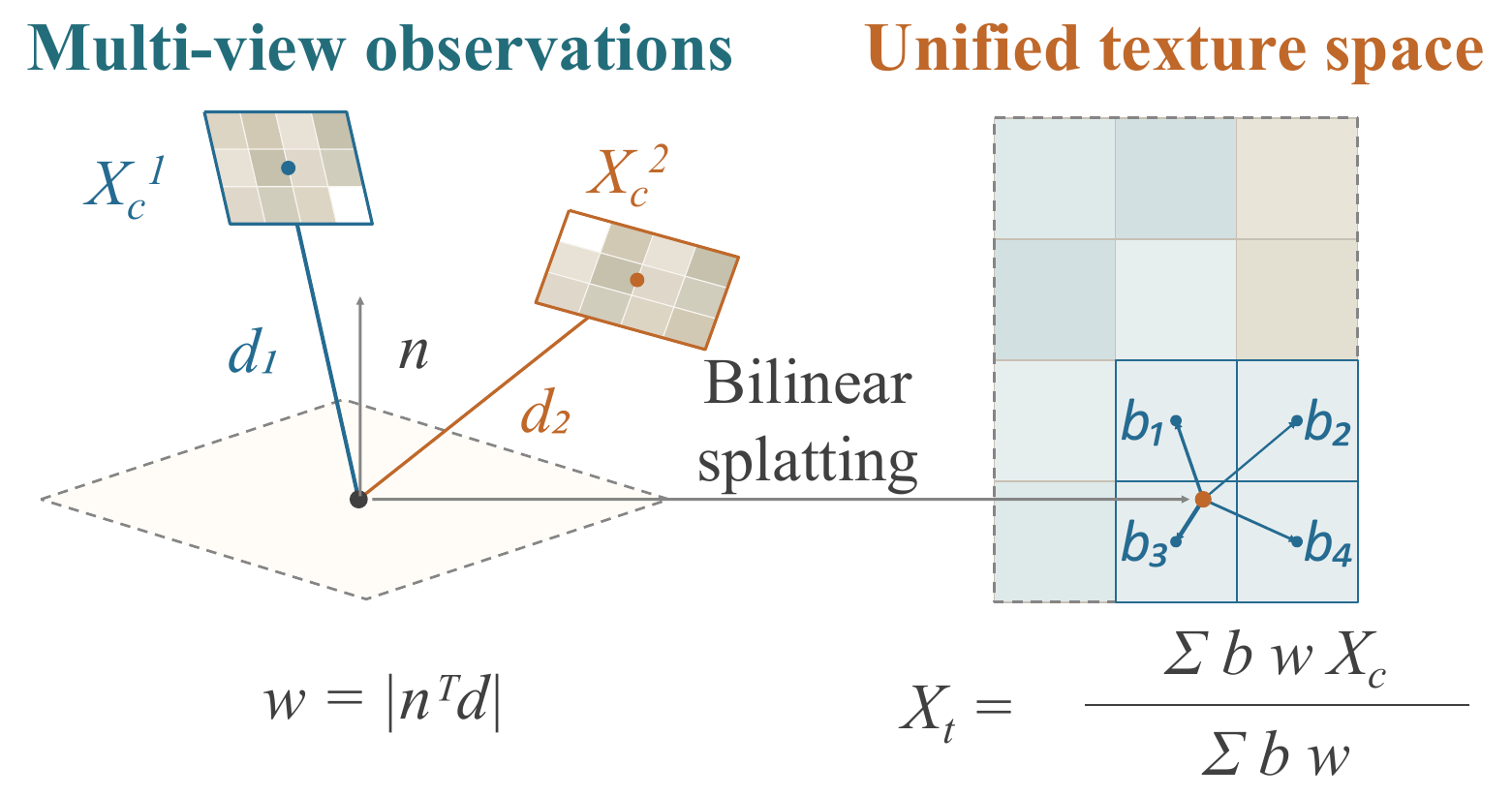}
    \caption{Weighted aggregation of multi-view observations into a unified texture space.}
    \label{fig:texture-aggregation}
\end{figure}

{As shown in Fig.~\ref{fig:texture-aggregation}, we aggregate the multi-view intrinsic observations $X_c$ into incomplete intrinsic texture maps $X_t$. Each observation is splatted onto the four neighboring texture pixels using bilinear weights. We additionally assign it a viewing-angle weight $w=|\mathbf{n}^{\top}\mathbf{d}|$, where $\mathbf{n}$ is the plane normal and $\mathbf{d}$ is the view direction, giving larger weight to more frontal observations. For each contributing $X_c$, we multiply its bilinear and viewing-angle weights to obtain a combined weight. The final value at each texture pixel is the weighted average of the observations. In this way, observations of the same plane location are merged at shared texture coordinates, while different views contribute complementary appearance information around the missing regions.}

\textbf{Inpainting in texture space.}
{We first construct a union inpainting mask $M_T$ to indicate the regions to be completed in texture space. For each object label $o\in\mathcal{O}{\mathrm{rm}}$, we select the corresponding Gaussian centers closest to the estimated plane and project them onto the texture grid to obtain the texture-space object mask $M_{\mathrm{obj}}^o$. We then combine the masks of all selected objects using elementwise logical OR to obtain mask $M_T$, i.e., $M_T=\bigvee_{o\in\mathcal{O}{\mathrm{rm}}} M_{\mathrm{obj}}^o$.} Then, we use LaMA~\citep{suvorov2022lama} as the 2D inpainting model $\Phi$ to complete the aggregated intrinsic textures. The diffuse radiance, Fresnel reflectance, and roughness maps are inpainted independently using the same mask $M_T$. For each intrinsic attribute, $\Phi$ takes the incomplete texture map and $M_T$ as input and predicts the missing regions, as: $\widehat{X_t}=(1-M_T)\odot X_t+M_T\odot\Phi(X_t,M_T),$ where $\odot$ denotes elementwise multiplication. This formulation preserves the input texture values outside the mask and uses the inpainted values within the selected regions.

\textbf{Reprojection to camera views.}
Using the camera-to-texture correspondence defined above, we sample the completed intrinsic textures to obtain supervision for each camera view:
\begin{equation}
X'_c(p)
=
\widehat{X_t}\!\left(\mathbf{u}_c(p)\right),
\label{eq:reprojected-target}
\end{equation}
Here, $X'_c$ denotes the reprojected intrinsic maps for camera $c$. We also map the texture-space mask $M_T$ to each camera view through the same correspondence to obtain the camera-view inpainting mask $M_c$. Camera pixels corresponding to the same plane location receive supervision from the same completed texture, reducing disagreement across views during subsequent Gaussian completion. Additional construction and sampling details are provided in Appendix~\ref{app:texture-details}.

\subsection{Geometrically Regularized Gaussian Completion}
\label{subsec:scene-completion-optimization}
The reprojected intrinsic maps provide supervision for reconstructing the previously occluded regions. However, appearance supervision alone does not uniquely determine the positions of the completion Gaussians. Direct optimization may therefore cause these Gaussians to drift along or away from the support surface, resulting in implausible geometry. To address this issue, we first lift the completed intrinsic textures into a 3D Gaussian representation on the estimated support plane, establishing a geometric reference for the missing regions. We then introduce geometric regularization to penalize deviations from this reference during optimization, while allowing local adjustments guided by the intrinsic supervision.

\textbf{Lifting to 3D.}
We construct completion Gaussians within the regions indicated by the texture-space inpainting mask $M_T$. Specifically, we sample texture coordinates $\mathbf{u}_i^0=(u_i^0,v_i^0)^{\top}$ where $M_T=1$ and lift them onto the estimated support plane. The inverse parameterization of this plane and the initial Gaussian centers are given by: $\mathbf{x}(u,v)=\mathbf{o}+u\mathbf{e}_u+v\mathbf{e}_v, \boldsymbol{\mu}_i^0=\mathbf{x}(u_i^0,v_i^0),$
where $\boldsymbol{\mu}_i^0$ denotes the initial center of the $i$-th completion Gaussian. This construction associates each Gaussian with a location in the completed texture domain. The initial tangential scales are determined by the sampling spacing. We initialize the diffuse radiance, Fresnel reflectance, and roughness of each Gaussian by sampling the completed intrinsic textures $\widehat{X_t}$ at $\mathbf{u}_i^0$. The resulting set of $N_h$ completion Gaussians, denoted by $\mathcal{G}_h$, provides an initial geometric configuration for the missing regions. Additional initialization details are provided in Appendix~\ref{app:gaussian-details}.

\textbf{Geometric regularization.}
We use the plane-based initialization as a geometric reference throughout optimization. Specifically, we introduce two soft regularization terms for the completion Gaussians:
\begin{equation}
\begin{aligned}
\mathcal{L}_{\mathrm{surf}}
&=
\sum_{i=1}^{N_h}
\left|
\mathbf{n}^{\top}\boldsymbol{\mu}_i+d
\right|,\\
\mathcal{L}_{\mathrm{anchor}}
&=
\sum_{i=1}^{N_h}
\left\|
T(\boldsymbol{\mu}_i)-\mathbf{u}_i^0
\right\|_1.
\end{aligned}
\label{eq:completion-geometric-losses}
\end{equation}
Here, $\boldsymbol{\mu}_i$ denotes the current center of the $i$-th completion Gaussian, and $\mathbf{u}_i^0$ denotes its initial texture coordinate. Since $\mathbf{n}$ is a unit normal, $\mathcal{L}_{\mathrm{surf}}$ penalizes the distance of each Gaussian center from the estimated support plane. The anchoring term $\mathcal{L}_{\mathrm{anchor}}$ penalizes displacement from the initial texture coordinates, discouraging excessive drift along the plane. Together, the surface and anchoring terms constrain normal and tangential displacement, respectively. Throughout completion, we keep $\mathcal{G}_{\mathrm{keep}}$, the original Gaussians representing the background and objects not selected for removal, fixed and optimize only $\mathcal{G}_h$ using the objective defined in Sec.~\ref{subsec:optimization}. After optimization, the edited scene is $\mathcal{G}_{\mathrm{edit}}=\mathcal{G}_{\mathrm{keep}}\cup\mathcal{G}_h$, combining the fixed original Gaussians with the optimized completion Gaussians for the previously occluded regions.

\subsection{Loss Functions}
\label{subsec:optimization}
We optimize the completion Gaussians using intrinsic supervision to recover their appearance and geometric regularization to constrain their spatial structure. The overall objective is
\begin{equation}
\begin{aligned}
\mathcal{L}
&=\mathcal{L}_{\mathrm{tex}}
+\lambda_{\mathrm{surf}}\mathcal{L}_{\mathrm{surf}}\\
&\quad+\lambda_{\mathrm{anchor}}\mathcal{L}_{\mathrm{anchor}},
\end{aligned}
\label{eq:completion-optimization-objective}
\end{equation}
where $\mathcal{L}_{\mathrm{tex}}$ provides intrinsic supervision, and the remaining terms regularize the geometry.

\textbf{Intrinsic supervision.}
For each training camera $c$, we render the combined scene $\mathcal{G}_{\mathrm{keep}}\cup\mathcal{G}_h$ to obtain intrinsic maps $X_{r,c}$. We supervise these maps using the reprojected targets $X'_c$ from Eq.~\ref{eq:reprojected-target} within the camera-view inpainting mask $M_c$:
\begin{equation}
\mathcal{L}_{\mathrm{tex}}
=
\sum_c
\left\|
M_c\odot
\left(
X_{r,c}-X'_c
\right)
\right\|_1,
\label{eq:intrinsic-completion-loss}
\end{equation}
where $\odot$ denotes elementwise multiplication, and the $L_1$ norm sums absolute differences over pixels and intrinsic channels. Since the targets are reprojected from the same completed texture maps, they provide shared intrinsic supervision across views.

\textbf{Geometric regularization.}
We use the two regularization terms defined in Eq.~\ref{eq:completion-geometric-losses}. The surface term $\mathcal{L}_{\mathrm{surf}}$ penalizes normal deviations from the estimated support plane, while the anchoring term $\mathcal{L}_{\mathrm{anchor}}$ penalizes tangential displacement from the initial texture coordinates. Their weights $\lambda_{\mathrm{surf}}$ and $\lambda_{\mathrm{anchor}}$ control the strength of these geometric constraints relative to the intrinsic supervision.


\section{Experiments}
\label{sec:result}

\newcommand{\expbest}[1]{\begingroup\setlength{\fboxsep}{1pt}\colorbox{red!18}{\strut #1}\endgroup}
\newcommand{\expsecond}[1]{\begingroup\setlength{\fboxsep}{1pt}\colorbox{yellow!35}{\strut #1}\endgroup}

\subsection{Implementation Details}

3DTexMOR takes multi-view images and their associated multi-object masks as input. We first optimize the initial 3DGS for 30k iterations. After removing the selected objects, we inpaint the intrinsic texture maps and optimize the completion Gaussians for 4k iterations. All experiments are conducted on an NVIDIA RTX 4090D GPU. Initial 3D Gaussian reconstruction takes approximately 40 minutes, increasing to about 2 hours when intrinsic decomposition is enabled. Gaussian segmentation requires an additional 10 minutes, while texture-space inpainting takes approximately 30 seconds. The subsequent Gaussian completion stage takes about 4 minutes. Additional implementation details are provided in the supplementary.

On Syn-Reflective, with approximately 30,000 Gaussians rendered at a resolution of $800\times800$, full physically-based rendering achieves 40.0~FPS on the RTX 4090D GPU. For faster rendering, the SH distillation strategy proposed in GOR-IS~\citep{zhao2026goris} can be further adopted, enabling efficient rendering with only a modest loss in rendering quality.

\subsection{Experimental Setup}

\textbf{Datasets.}
We evaluate 3DTexMOR on both synthetic and real-world scenes. The synthetic benchmark contains two datasets. \textit{Syn-Reflective} consists of three reflective scenes from GOR-IS~\citep{zhao2026goris}. \textit{Syn-Diffuse} contains three diffuse scenes with dense object layouts, providing a more challenging setting for multi-object removal. These scenes are generated using Blender. To evaluate generalization, we further construct two real-world datasets from public datasets~\citep{kerr2023lerf,wang2026inpaint360gs,barron2022mipnerf360}. \textit{Real-Single} contains three single-object scenes, while \textit{Real-Multiple} contains four multi-object scenes. For all scenes, we use SAM2~\citep{ravi2024sam2} to obtain view-associated object masks. We quantitatively evaluate multi-object removal on synthetic scenes, reporting diffuse and reflective scenes separately. For real-world scenes, quantitative evaluation is unavailable due to the lack of ground truth.

\begin{table*}[t]
  \caption{Quantitative comparison on the Syn-Diffuse and the Syn-Reflective datasets. We report full-image metrics and mask-domain metrics. Higher PSNR/SSIM and lower LPIPS/FID are better. \expbest{Red} highlights the best result and \expsecond{yellow} highlights the second-best result in each group.}
  \label{tab:main-comparison}
  \centering
  \small
  \setlength{\tabcolsep}{1.5pt}
  \renewcommand{\arraystretch}{1.15}
  \resizebox{\linewidth}{!}{%
  \begin{tabular}{@{}lcccc@{\hspace{8pt}}cccc@{}}
    \toprule
    & \multicolumn{4}{c}{\textit{Syn-Diffuse dataset}}
    & \multicolumn{4}{c}{\textit{Syn-Reflective dataset}} \\
    \cmidrule(lr){2-5}\cmidrule(lr){6-9}
    Method & PSNR / mPSNR $\uparrow$ & SSIM / mSSIM $\uparrow$ &
    LPIPS / mLPIPS $\downarrow$ & FID / mFID $\downarrow$ &
    PSNR / mPSNR $\uparrow$ & SSIM / mSSIM $\uparrow$ &
    LPIPS / mLPIPS $\downarrow$ & FID / mFID $\downarrow$ \\
    \midrule
    Ours & \expbest{31.31 / 28.79} & \expbest{0.932 / 0.935} & \expbest{0.096 / 0.098} & \expbest{34.4 / 76.5} & \expbest{30.20 / 29.34} & \expbest{0.984 / 0.981} & \expbest{0.089 / 0.103} & \expbest{71.4 / 81.3} \\
    GOR-IS & \expsecond{25.04 / 20.94} & \expsecond{0.878 / 0.910} &
    \expsecond{0.232 / 0.179} & \expsecond{199.0 / 142.9} &
    \expsecond{28.90 / 26.81} & \expsecond{0.973 / 0.979} &
    \expsecond{0.118 / 0.109} & \expsecond{76.8 / 87.9} \\
    GPGS & 21.18 / 16.67 & 0.821 / 0.873 & 0.268 / 0.202 & 228.4 / 171.6 &
    22.17 / 17.54 & 0.858 / 0.868 & 0.257 / 0.192 & 217.6 / 164.2 \\
    3DGIC & {22.49 / 17.50} & 0.805 / 0.867 &
    0.289 / 0.198 & {211.7 / 162.8} &
    23.61 / 18.38 & 0.845 / 0.864 &
    0.275 / 0.188 & 201.1 / 154.7 \\
    AuraFusion360 & 20.44 / 16.92 & 0.799 / 0.866 &
    0.290 / 0.210 & 256.5 / 180.3 &
    21.46 / 17.77 & 0.839 / 0.845 &
    0.276 / 0.200 & 243.7 / 171.3 \\
    Inpaint360GS & 20.27 / 15.80 & {0.845 / 0.885} &
    {0.238 / 0.195} & 212.5 / 174.4 &
    21.28 / 16.59 & {0.887 / 0.891} &
    {0.226 / 0.185} & 201.9 / 165.7 \\
    \bottomrule
  \end{tabular}}
\end{table*}

\begin{figure*}[t]
  \centering
\includegraphics[width=\textwidth,height=0.56\textheight,keepaspectratio]{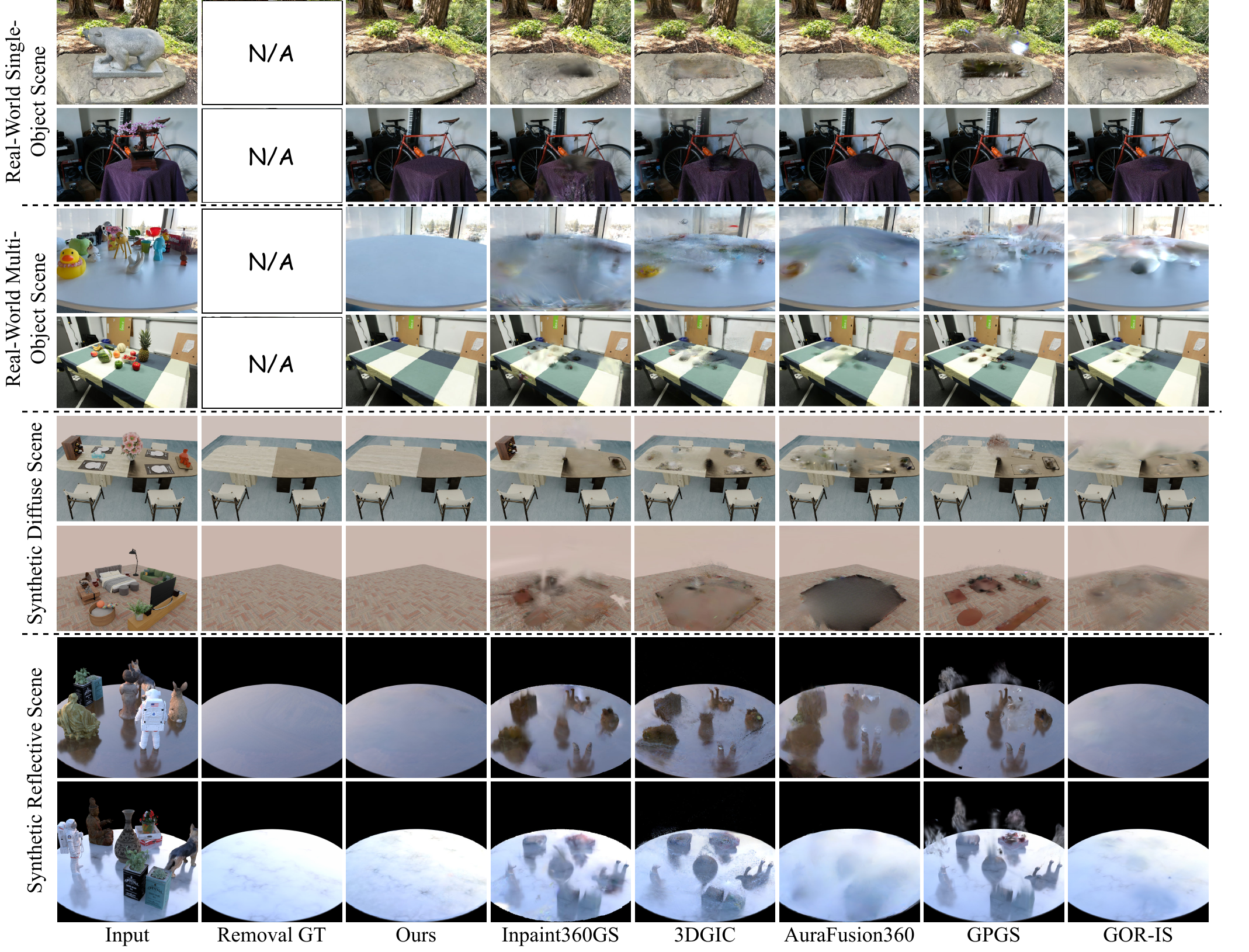}
  \caption{
  {Visual comparisons with prior works on real-world and synthetic scenes, showing 3DTexMOR produces more consistent completion, with fewer artifacts and less blurriness.}
}
  \label{fig:synthetic-results}
\end{figure*}%

\textbf{Baselines.}
We compare 3DTexMOR with five recent 3DGS-based object removal methods: 3DGIC~\citep{huang2025_3dgic}, AuraFusion360~\citep{wu2025aurafusion360}, Inpaint360GS~\citep{wang2026inpaint360gs}, GPGS~\citep{lee2026gpgs}, and GOR-IS~\citep{zhao2026goris}. We use the same inputs and train/test splits for all methods. Since some baselines support only single-object removal, we adapt their implementations to handle multi-object scenes by removing the target objects iteratively.

\textbf{Metrics.}
We report PSNR, SSIM~\citep{wang2004ssim}, LPIPS~\citep{zhang2018lpips} with AlexNet, and FID~\citep{heusel2017fid} with Inception~V3. PSNR and SSIM measure pixel-level fidelity, LPIPS measures perceptual similarity, and FID evaluates the distributional realism of rendered novel views. Since object removal mainly affects the regions originally occupied by the removed objects, we additionally report mask-domain metrics, denoted as mPSNR, mSSIM, mLPIPS, and mFID. Following GOR-IS, mask-domain evaluation uses the bounding rectangle enclosing the union of all removal masks.

\subsection{Comparisons with prior works}

We compare 3DTexMOR with several baselines on four datasets spanning real and synthetic scenes, single- and multi-object removal settings, and diffuse and specular surfaces. The visual comparisons in Fig.~\ref{fig:synthetic-results} show that the baselines produce noticeable artifacts, whereas 3DTexMOR yields fewer artifacts and more closely matches the GT appearance. The advantage is particularly clear in complex multi-object scenes, where individual views provide limited visual context for completing the occluded regions.


Table~\ref{tab:main-comparison} reports quantitative results on Syn-Diffuse and Syn-Reflective. 3DTexMOR achieves the best full-image and mask-domain metrics on both datasets. Compared with GOR-IS, full-image PSNR increases by 6.27~dB on Syn-Diffuse and 1.30~dB on Syn-Reflective, while full-image LPIPS decreases by 58.6\% and 24.6\%, respectively. Together with the qualitative comparisons, these results demonstrate the effectiveness of 3DTexMOR across diffuse and reflective scenes.

\raggedbottom
\subsection{Ablation Studies}

\begin{figure*}[t]
  \centering
  \includegraphics[width=0.85\linewidth]{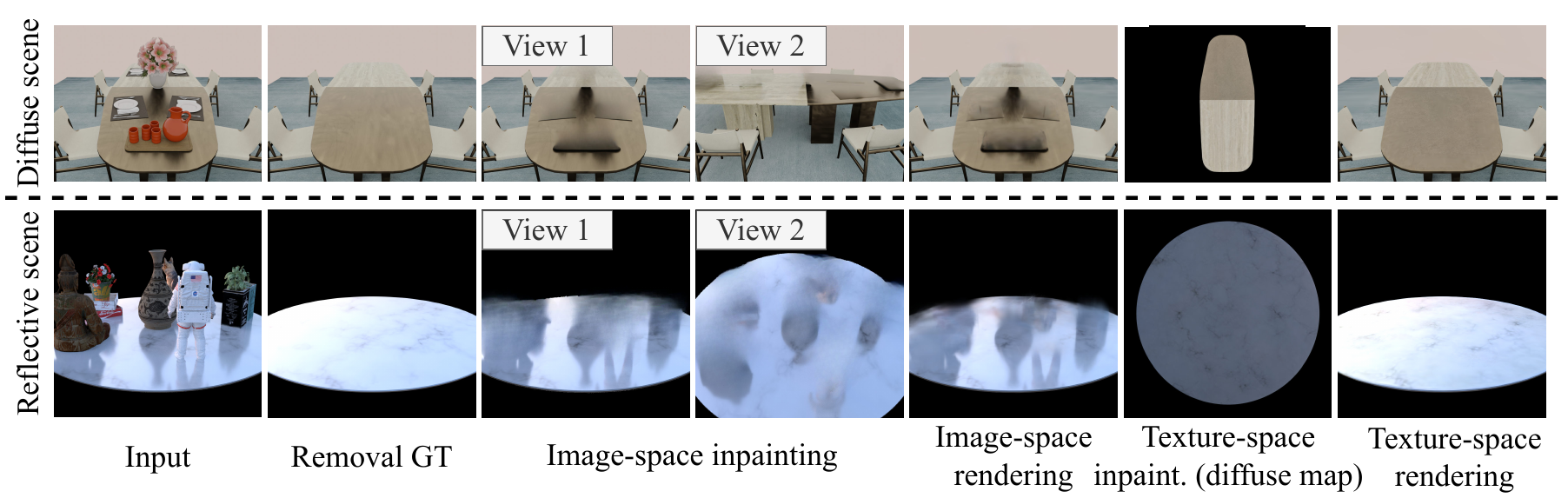}
  \caption{{Ablation of texture-space inpainting. Compared with image-space inpainting, completing a unified texture space yields cleaner inpainted results and renderings with fewer artifacts.}}
  \label{fig:texture-vs-screen}
\end{figure*}

\noindent\textbf{Texture-space inpainting.}
We evaluate the contribution of texture-space inpainting by replacing it with image-space inpainting, which applies LaMa independently to multiple camera views, while keeping all other components unchanged. We compare the two approaches on both the Syn-Diffuse and Syn-Reflective datasets. As shown in Fig.~\ref{fig:texture-vs-screen}, image-space inpainting produces noticeable artifacts in previously occluded regions, which propagate to the rendered results. In contrast, our method performs inpainting in a unified texture space and reprojects the completed maps to provide shared supervision across views, producing cleaner and more consistent appearance in the completed regions. Table~\ref{tab:texture-vs-screen} reports PSNR gains of 5.58~dB on Syn-Diffuse and 8.59~dB on Syn-Reflective over image-space inpainting. Additional qualitative results can be found in Fig.~\ref{fig:single-vs-multi} of the supplementary material.



\begin{figure*}[t]
  \centering
  \includegraphics[width=0.85\linewidth]{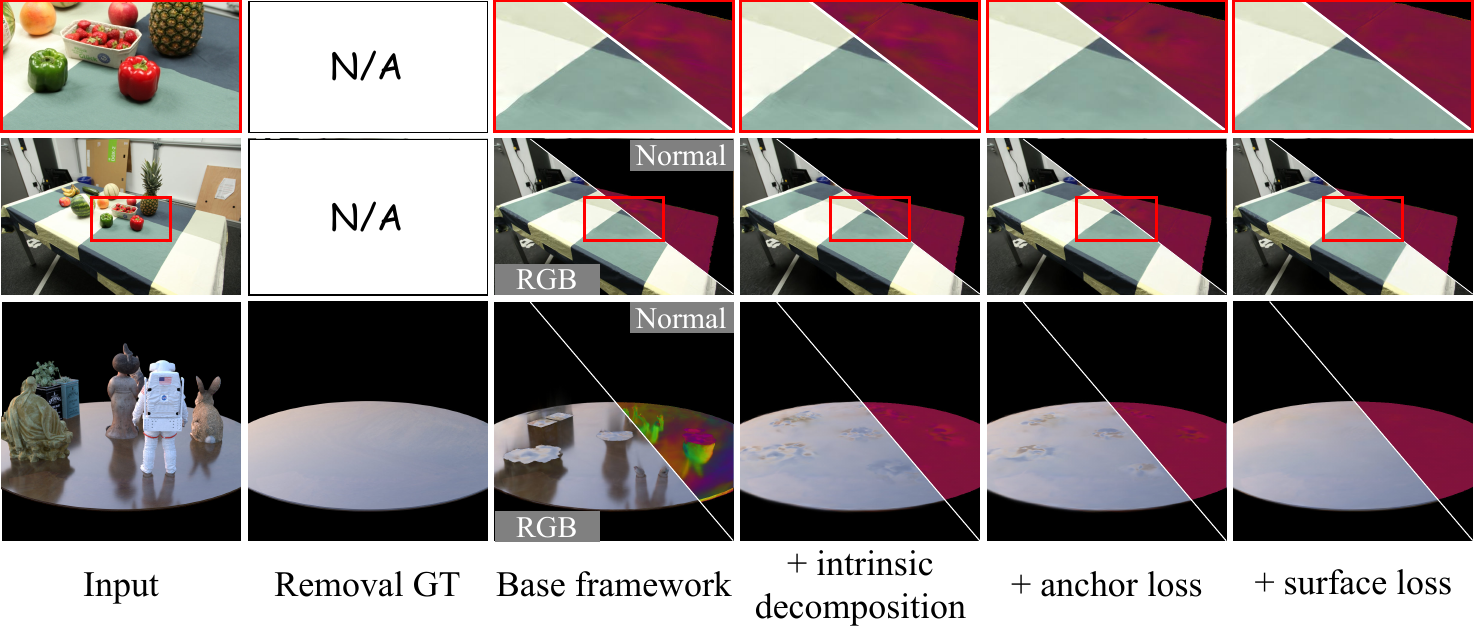}
  \caption{{Ablation of intrinsic decomposition and geometric regularization. Starting from the texture-space-only baseline, we progressively add intrinsic decomposition, the anchor loss, and the surface loss to reduce reflection artifacts and improve completion geometry.}}
  \label{fig:inpainting-modules}
\end{figure*}

\noindent\textbf{Intrinsic decomposition.}
Starting from a base framework with only texture-space inpainting enabled, we add view-independent intrinsic decomposition. On reflective scenes, this removes reflections of the deleted objects that otherwise remain in the completed regions. Fig.~\ref{fig:inpainting-modules} shows the disappearance of these reflections in the RGB renderings and a reduction in the associated geometry artifacts in the normal maps. Table~\ref{tab:inpainting-modules} shows that intrinsic decomposition improves all reported metrics on Syn-Reflective, increasing PSNR from 21.38 to 28.58~dB. On Syn-Diffuse, the metrics remain unchanged, consistent with the absence of strong view-dependent reflections.

\noindent\textbf{Geometric Regularization.}
The anchor loss discourages completion Gaussians from drifting along the estimated plane by keeping their texture coordinates close to their initial positions. In Fig.~\ref{fig:inpainting-modules}, adding this loss reduces the area affected by geometric irregularities, which is particularly clear in the normal maps. The surface loss further penalizes the perpendicular distance between Gaussian centers and the estimated plane. Adding this term produces more uniform normals and visibly improves the RGB renderings, indicating flatter geometry in the completed regions. Together, the two terms constrain drift along and away from the plane. Table~\ref{tab:inpainting-modules} further shows that the performance improves progressively as each component is introduced, with the full model achieving the best metrics.

\begin{figure*}[t]
  \captionsetup{type=table}
  \caption{Ablation of image-space versus texture-space inpainting on Syn-Diffuse and Syn-Reflective, reported separately. Each pair gives full-image / mask-domain metrics. The visual comparison is shown in Fig.~\ref{fig:texture-vs-screen}. Red highlights the better result.}
  \label{tab:texture-vs-screen}
  \centering
  \footnotesize
  \setlength{\tabcolsep}{1.5pt}
  \renewcommand{\arraystretch}{1.12}
  \resizebox{\linewidth}{!}{%
  \begin{tabular}{@{}lcccc@{\hspace{8pt}}cccc@{}}
    \toprule
    & \multicolumn{4}{c}{\textit{Syn-Diffuse dataset}}
    & \multicolumn{4}{c}{\textit{Syn-Reflective dataset}} \\
    \cmidrule(lr){2-5}\cmidrule(lr){6-9}
    Variant & PSNR / mPSNR $\uparrow$ & SSIM / mSSIM $\uparrow$ &
    LPIPS / mLPIPS $\downarrow$ & FID / mFID $\downarrow$ &
    PSNR / mPSNR $\uparrow$ & SSIM / mSSIM $\uparrow$ &
    LPIPS / mLPIPS $\downarrow$ & FID / mFID $\downarrow$ \\
    \midrule
    Image-space & 25.73 / 22.47 & 0.917 / 0.912 & 0.142 / 0.167 & 105.3 / 145.4 & 21.61 / 18.48 & 0.927 / 0.903 & 0.154 / 0.200 & 207.7 / 219.2 \\
    Texture-space & \expbest{31.31 / 28.79} & \expbest{0.932 / 0.935} & \expbest{0.096 / 0.098} & \expbest{34.4 / 76.5} & \expbest{30.20 / 29.34} & \expbest{0.984 / 0.981} & \expbest{0.089 / 0.103} & \expbest{71.4 / 81.3} \\
    \bottomrule
  \end{tabular}}
\end{figure*}

\begin{table*}[t]
  \caption{Component ablation with texture-space inpainting enabled. Intrinsic decomposition, the anchor loss, and the surface loss are added cumulatively. Each pair gives full-image / mask-domain metrics. Red and yellow highlight the best and second-best results, respectively.}
  \label{tab:inpainting-modules}
  \centering
  \footnotesize
  \setlength{\tabcolsep}{1.5pt}
  \renewcommand{\arraystretch}{1.12}
  \resizebox{\linewidth}{!}{%
  \begin{tabular}{@{}lcccc@{\hspace{8pt}}cccc@{}}
    \toprule
    & \multicolumn{4}{c}{\textit{Syn-Diffuse dataset}}
    & \multicolumn{4}{c}{\textit{Syn-Reflective dataset}} \\
    \cmidrule(lr){2-5}\cmidrule(lr){6-9}
    Variant & PSNR / mPSNR $\uparrow$ & SSIM / mSSIM $\uparrow$ &
    LPIPS / mLPIPS $\downarrow$ & FID / mFID $\downarrow$ &
    PSNR / mPSNR $\uparrow$ & SSIM / mSSIM $\uparrow$ &
    LPIPS / mLPIPS $\downarrow$ & FID / mFID $\downarrow$ \\
    \midrule
    Base framework & 30.03 / 27.04 & 0.925 / 0.910 & 0.123 / 0.132 & 51.8 / 144.3 & 21.38 / 18.23 & 0.915 / 0.887 & 0.190 / 0.242 & 234.3 / 253.9 \\
    + Intrinsic decomposition & 30.03 / 27.04 & 0.925 / 0.910 & 0.123 / 0.132 & 51.8 / 144.3 & 28.58 / 27.83 & 0.948 / 0.938 & 0.107 / 0.129 & 141.9 / 142.2 \\
    + Anchor loss & \expsecond{30.50 / 28.36} & \expsecond{0.927 / 0.924} & \expsecond{0.106 / 0.109} & \expsecond{44.9 / 121.1} & \expsecond{29.08 / 28.46} & \expsecond{0.953 / 0.945} & \expsecond{0.094 / 0.112} & \expsecond{114.7 / 125.4} \\
    + Surface loss & \expbest{31.31 / 28.79} & \expbest{0.932 / 0.935} & \expbest{0.096 / 0.098} & \expbest{34.4 / 76.5} & \expbest{30.20 / 29.34} & \expbest{0.984 / 0.981} & \expbest{0.089 / 0.103} & \expbest{71.4 / 81.3} \\
    \bottomrule
  \end{tabular}}
\end{table*}

\section{Conclusion}
\label{sec:conclu}

In this paper, we propose 3DTexMOR, a novel framework for high-quality 3D multi-object removal. By aggregating complementary multi-view observations in a unified texture space shared across camera views, 3DTexMOR provides richer visual information for recovering occluded regions while reducing appearance inconsistencies. We further introduce a geometrically regularized Gaussian completion module that complements texture supervision with geometric constraints to recover plausible 3D structure in the completed regions. Experiments on synthetic and real-world scenes demonstrate that 3DTexMOR outperforms recent Gaussian object removal methods.

\textit{Limitations and future work.}
3DTexMOR assumes that the completion regions can be approximated by a common plane. This assumption enables stable texture-space construction but limits the method on curved, deformable, or irregular surfaces, where planar mapping may misalign the reprojected supervision. Detailed experiments on complex backgrounds are provided in Sec.~\ref{app:multi-complex-surface-limitations} of the supplementary material. Extending the texture-space representation to more general surfaces is an important direction for future work.

\clearpage
\makeatletter
\global\@colht=\textheight
\global\@colroom=\textheight
\global\vsize=\textheight
\makeatother
\section*{LLM Usage Statement}
OpenAI Codex was used to assist with manuscript refinement and to review the clarity and consistency of the method description and mathematical notation. It also assisted with literature searches and reference organization, LaTeX editing and formatting, experimental environment setup, code review, debugging and optimization, experiment deployment and execution, and the organization of experimental results. The authors reviewed the AI-assisted revisions and code changes and remain fully responsible for the methods, experimental results, interpretations, and claims reported in this paper.

\clearpage
\appendix
\numberwithin{equation}{section}
\numberwithin{figure}{section}
\section{Technical Details}
\label{app:method-details}

This appendix expands the representation and two completion modules in
Secs.~\ref{subsec:object-intrinsic-decomposition}--\ref{subsec:scene-completion-optimization},
together with the objective in Sec.~\ref{subsec:optimization}.
We first detail the intrinsic attributes and reflection renderer, then
describe texture-space construction, inpainting, and reprojection.
Finally, we specify Gaussian initialization and explain how intrinsic
supervision and geometric regularization guide completion.

\subsection{Intrinsic Decomposition and Reflection Rendering}
\label{app:reflection-details}

\paragraph{Gaussian attributes and deferred maps.}
We use the reflection model of GOR-IS~\citep{zhao2026goris} to separate the
diffuse and glossy terms in Sec.~\ref{subsec:object-intrinsic-decomposition}.
Each Gaussian carries diffuse radiance $\mathbf{d}_i$, normal-incidence Fresnel
reflectance $\mathbf{f}_i$, and roughness $r_i$.
Deferred rasterization produces their per-view maps $D_c$, $F_c$, and
$R_c$, together with a normal map $N_c$ from the geometry branch.
The diffuse term is view-independent radiance under the fixed capture
lighting and already includes diffuse illumination; it is not an estimate
of unlit albedo.
The remaining glossy component is evaluated from the viewing direction,
surface geometry, intrinsic attributes, and incident radiance.

\paragraph{Reflected rays and incident radiance.}
At pixel $p$, let $\mathbf{n}_s$ be the unit shading normal from $N_c$.
This local normal is distinct from the fitted plane normal
$\mathbf{n}$ used to construct texture space.
Let $\mathbf{d}_{c,p}$ be the unit ray direction from camera $c$ to the
shading point at pixel $p$.
The reflected direction is
\begin{equation}
  \boldsymbol{\omega}_r
  =\mathbf{d}_{c,p}
   -2(\mathbf{d}_{c,p}^{\top}\mathbf{n}_s)\mathbf{n}_s,
  \qquad \boldsymbol{\omega}_o=-\mathbf{d}_{c,p}.
  \label{eq:reflected-ray-direction}
\end{equation}
We trace a secondary ray from the shading point along
$\boldsymbol{\omega}_r$ through the Gaussian scene to obtain the near-field
radiance.
An optimizable environment map provides the far-field contribution according
to ray visibility.
Together these contributions form the incident-radiance map $L_{i,c}$.
Since the near-field term queries the scene itself, removing an object can
change the reflected radiance at a different surface location.

\paragraph{Fresnel response and roughness.}
The renderer converts $F_c$ into the angle-dependent Fresnel factor
$\mathcal{F}_c$ using Schlick's approximation.
It forms an ideal reflection map and applies a roughness-guided screen-space
filter $\mathcal{H}$ to approximate glossy reflection:
\begin{equation}
  S_c=\mathcal{F}_c\odot L_{i,c},
  \qquad G_c=\mathcal{H}(S_c,R_c).
  \label{eq:glossy-reflection-filter}
\end{equation}
The intrinsic Fresnel reflectance $F_c$ represents a view-independent
attribute, whereas $\mathcal{F}_c$ depends on the viewing direction.
Only the intrinsic attribute enters texture space; the angle-dependent
factor is evaluated during reflection rendering.

\paragraph{Intrinsic maps as completion inputs.}
The decomposition separates the view-independent attributes in $X_c$
from the view-dependent glossy component $G_c$.
Texture-space inpainting operates on the former, so observations can be
aggregated across views without directly combining their highlights and
reflections. The glossy component is evaluated from the edited scene,
as described in Appendix~\ref{app:gaussian-details}.

\subsection{Texture-Space Inpainting Details}
\label{app:texture-details}

\paragraph{Plane estimation.}
Let $\mathcal{P}_s=\{\mathbf{x}_i\}_{i=1}^{N_s}$ contain the background
Gaussian centers used to estimate the plane in
Sec.~\ref{subsec:reversible-texture-mapping}.
The least-squares fit minimizes their squared distances to a plane:
\begin{equation}
  \min_{\mathbf{n},d}\sum_{i=1}^{N_s}
    (\mathbf{n}^{\top}\mathbf{x}_i+d)^2,
  \qquad \|\mathbf{n}\|_2=1.
  \label{eq:texture-support-plane}
\end{equation}
The unit-normal constraint makes
$|\mathbf{n}^{\top}\mathbf{x}+d|$ the point-to-plane distance in world units.
The fitted plane provides the common geometric reference for texture
projection, object masks, Gaussian initialization, and surface regularization.

\paragraph{Plane coordinates and the texel grid.}
We choose an origin $\mathbf{o}$ on the fitted plane and two orthonormal
tangent axes satisfying $\mathbf{e}_u\times\mathbf{e}_v=\mathbf{n}$.
The mapping in Sec.~\ref{subsec:reversible-texture-mapping} assigns each
3D point two tangential coordinates. Its inverse parameterization in
Sec.~\ref{subsec:scene-completion-optimization} lifts these coordinates
onto the plane.
In particular, $T(\mathbf{x}(u,v))=(u,v)^{\top}$.
For a point away from the plane, $T$ still gives its tangential coordinates;
its normal displacement is measured separately by
$\mathbf{n}^{\top}\mathbf{x}+d$.
This distinction is used by the two position penalties in
Eq.~(\ref{eq:completion-geometric-losses}).

The texture domain $\Omega$ covers the projected background observations and
the footprints of all selected objects.
We discretize $\Omega$ into a regular grid with texel spacings
$\Delta_u$ and $\Delta_v$, measured in world units along the tangent axes.
To make the sampling convention explicit, let
$\mathbf{t}_{\mathrm{ref}}=(u_{\mathrm{ref}},v_{\mathrm{ref}})$ be the
texture coordinate of the first texel center.
The conversion to continuous grid indices is
\begin{equation}
  \gamma(u,v)
  =\left(\frac{u-u_{\mathrm{ref}}}{\Delta_u},
          \frac{v-v_{\mathrm{ref}}}{\Delta_v}\right)^{\top}.
  \label{eq:texture-grid-coordinates}
\end{equation}
Integer indices identify texel centers.
The geometric mapping $T$ and the anchor loss use plane coordinates, while
splatting, mask rasterization, and the sampling stride use grid coordinates.
Thus, a stride of $s_T$ texels corresponds to distances
$s_T\Delta_u$ and $s_T\Delta_v$ on the surface.

\paragraph{Camera-to-texture correspondence.}
For pixel $p$ in camera $c$, the calibrated camera parameters define the
world-space ray $\mathbf{c}_c+t\mathbf{d}_{c,p}$, where $\mathbf{c}_c$
is the camera center and $\mathbf{d}_{c,p}$ is a unit direction pointing
from the camera into the scene.
Substituting the ray into the plane equation yields
\begin{equation}
  \begin{aligned}
    t_c(p)&=-\frac{\mathbf{n}^{\top}\mathbf{c}_c+d}
                        {\mathbf{n}^{\top}\mathbf{d}_{c,p}},\\
    \mathbf{x}_c(p)&=\mathbf{c}_c+t_c(p)\mathbf{d}_{c,p},
    \qquad \mathbf{u}_c(p)=T(\mathbf{x}_c(p)).
  \end{aligned}
  \label{eq:texture-ray-plane-intersection}
\end{equation}
A geometric correspondence is valid only if the ray is not parallel to the
plane, $t_c(p)>0$, and $\mathbf{u}_c(p)\in\Omega$.
These conditions ensure a forward intersection inside the texture extent;
they do not by themselves establish that the background is observed.
For aggregation, we additionally require a visible background
observation and exclude pixels occluded by retained objects.
We denote the resulting set of camera--pixel pairs by $\mathcal{V}$.

\paragraph{Multi-view observation aggregation.}
Let $X_{c,q}(p)$ denote attribute $q\in\mathcal{Q}=\{D,F,R\}$ at
pixel $p$ in the intrinsic maps $X_c=\{D_c,F_c,R_c\}$.
We aggregate each attribute using the same camera-to-texture correspondence.
An observation at $\mathbf{u}_c(p)$ contributes to its four neighboring texels by
bilinear splatting.
For a texel centered at plane coordinate $\mathbf{z}$, its coefficient is
\begin{equation}
  b(\mathbf{z},\mathbf{t})
  =\prod_{k\in\{u,v\}}
    \max\!\left(0,1-
      |\gamma_k(\mathbf{z})-\gamma_k(\mathbf{t})|\right).
  \label{eq:texture-bilinear-weight}
\end{equation}
Only neighboring texels inside the grid are accumulated.
We also weight each observation by
$w_{c,p}=|\mathbf{n}^{\top}\mathbf{d}_{c,p}|$.
Because both vectors are normalized, this weight lies in $[0,1]$ and
decreases for grazing views.
The corresponding incomplete intrinsic texture is
\begin{equation}
  X_{t,q}(\mathbf{z})=
  \frac{\sum_{(c,p)\in\mathcal{V}}
    b(\mathbf{z},\mathbf{u}_c(p))\,w_{c,p}\,X_{c,q}(p)}
       {\sum_{(c,p)\in\mathcal{V}}
    b(\mathbf{z},\mathbf{u}_c(p))\,w_{c,p}+\epsilon},
  \label{eq:texture-observation-aggregation}
\end{equation}
where $\epsilon>0$ stabilizes the denominator and the same scalar weights
apply to each component of a vector-valued channel.
The denominator measures accumulated observation support; adding $\epsilon$
does not turn a texel without observations into an observed surface sample.
Equation~\eqref{eq:texture-observation-aggregation} is applied separately to
the three channels specified in Appendix~\ref{app:reflection-details}.

\paragraph{Object footprints and the completion mask.}
We retain the target objects' Gaussian centers before deleting their
Gaussians from the scene.
For object $o\in\mathcal{O}_{\mathrm{rm}}$, let
$\mathcal{P}_o=\{\mathbf{x}_i^o\}$ be these centers and
$\delta_i^o=|\mathbf{n}^{\top}\mathbf{x}_i^o+d|$ their distances to the
estimated plane.
As in Sec.~\ref{subsec:reversible-texture-mapping}, we select the centers
closest to the plane and project them using $T$ to obtain the object mask
$M_{\mathrm{obj}}^o$ on the texture grid.
Selecting centers by their point-to-plane distances associates the mask
with the object's footprint on the estimated plane.

The joint mask is the pixelwise union
\begin{equation}
  M_T=\bigvee_{o\in\mathcal{O}_{\mathrm{rm}}}M_{\mathrm{obj}}^o.
  \label{eq:texture-footprint-mask}
\end{equation}
We use this footprint union directly as the completion mask.
The observation-validity set $\mathcal{V}$ determines which values enter the
texture, whereas $M_T$ determines which texture values are regenerated.
These two roles are distinct; observed texels are not subtracted from the
footprint union.

\paragraph{Intrinsic aggregation and completion.}
After deleting the selected objects, we rasterize the retained scene to
obtain $X_c$ and aggregate these observations into the incomplete intrinsic
textures $X_t=\{X_{t,D},X_{t,F},X_{t,R}\}$ using
Eq.~\eqref{eq:texture-observation-aggregation}.
We apply LaMA independently to the three textures with the same mask $M_T$,
following the inpainting formulation in Sec.~\ref{subsec:reversible-texture-mapping}.
This produces the completed maps $\widehat{X}_{t,D}$,
$\widehat{X}_{t,F}$, and $\widehat{X}_{t,R}$, collectively denoted by
$\widehat{X_t}$ in the main text.
Generated values are retained inside $M_T$, and the aggregated input remains
unchanged outside it.
All three textures therefore refer to the same physical surface locations.
Normals remain in the geometry branch and are not inpainted as texture
values.
The glossy term $G$ is also excluded from inpainting because it depends on
the current scene and viewing direction.

\paragraph{Geometric reprojection.}
We reproject the completed intrinsic maps by sampling them at the
camera-to-texture coordinates $\mathbf{u}_c(p)$, as in
Eq.~(\ref{eq:reprojected-target}).
Projection and reprojection use the same fitted plane and coordinate frame,
so pixels corresponding to the same plane location share a completed
texture value across views.
For target construction, the completed surface must be in front of the
camera, inside the texture domain, and unoccluded by retained content.
Unlike an aggregation sample, such a target need not have an original
background observation: it may lie in the previously hidden region.

\paragraph{Reprojected targets and the supervision mask.}
We convert the completed textures into per-view targets by sampling
\begin{equation}
  X'_{c,q}(p)=\widehat{X}_{t,q}\!\left(\mathbf{u}_c(p)\right),
  \qquad q\in\mathcal{Q}.
  \label{eq:intrinsic-reprojection-target}
\end{equation}
Let $v_c(p)$ indicate a valid plane intersection that is unoccluded by
retained scene content.
The completion-supervision mask can be written as
\begin{equation}
  M_c(p)=
  \begin{cases}
    M_T(\mathbf{u}_c(p)), & v_c(p)=1,\\
    0, & v_c(p)=0.
  \end{cases}
  \label{eq:intrinsic-supervision-mask}
\end{equation}
Thus, a retained foreground object does not receive a completed background
target merely because its camera ray intersects the plane behind it.

\subsection{Gaussian Initialization and Optimization Details}
\label{app:gaussian-details}

\paragraph{Sampling the initialization region.}
We initialize completion Gaussians within the same mask $M_T$ used for
texture-space inpainting.
Let $\mathcal{S}_{s_T}$ denote grid locations sampled with stride $s_T$
texels and expressed in plane coordinates. We retain only samples inside
$M_T$:
\begin{equation}
  \mathcal{U}_h
  =\{\mathbf{u}\in\mathcal{S}_{s_T}\mid M_T(\mathbf{u})=1\}
  =\{\mathbf{u}_i^0=(u_i^0,v_i^0)^{\top}\}_{i=1}^{N_h}.
  \label{eq:completion-texture-samples}
\end{equation}
The superscript $0$ denotes initialization. Each selected coordinate
supplies both the initial center and the intrinsic attributes of a
completion Gaussian.

\paragraph{Centers and scales.}
Each sampled location is lifted onto the estimated plane:
\begin{equation}
  \boldsymbol{\mu}_i^0
  =\mathbf{x}(u_i^0,v_i^0)
  =\mathbf{o}+u_i^0\mathbf{e}_u+v_i^0\mathbf{e}_v.
  \label{eq:completion-center-initialization}
\end{equation}
It therefore satisfies $\mathbf{n}^{\top}\boldsymbol{\mu}_i^0+d=0$ and
$T(\boldsymbol{\mu}_i^0)=\mathbf{u}_i^0$ at initialization.
This correspondence also identifies the texture location from which its
appearance is sampled.

As in Sec.~\ref{subsec:scene-completion-optimization}, the initial
tangential scales are determined by the sampling spacing.
In the texture coordinates defined above, this spacing corresponds to
$s_T\Delta_u$ and $s_T\Delta_v$ in world units.
We denote the initial scale vector by $\mathbf{s}_i^0$.

\paragraph{Intrinsic attributes.}
We query the completed texture maps at the same initial coordinate:
\begin{equation}
  \mathbf{d}_i^0=\widehat{X}_{t,D}(\mathbf{u}_i^0),
  \quad
  \mathbf{f}_i^0=\widehat{X}_{t,F}(\mathbf{u}_i^0),
  \quad
  r_i^0=\widehat{X}_{t,R}(\mathbf{u}_i^0).
  \label{eq:completion-attribute-initialization}
\end{equation}
These attributes provide the initial appearance of $\mathcal{G}_h$.
Combining the completion Gaussians with the retained scene gives
$\mathcal{G}_{\mathrm{keep}}\cup\mathcal{G}_h$, whose rendered intrinsic
maps are supervised during completion.

\paragraph{Intrinsic supervision.}
The masked $L_1$ loss in Eq.~(\ref{eq:intrinsic-completion-loss}) compares
the rendered intrinsic maps $X_{r,c}$ with the fixed reprojected targets
$X'_c$ inside the camera-view inpainting mask $M_c$.
Writing the attribute channels explicitly gives
\begin{equation}
  \mathcal{L}_{\mathrm{tex}}
  =\sum_c\sum_{q\in\mathcal{Q}}
    \left\|M_c\odot\left(X_{r,c,q}-X'_{c,q}\right)\right\|_1.
  \label{eq:appendix-intrinsic-loss}
\end{equation}
Here, $X_{r,c,q}$ denotes channel $q$ of the current rendered maps.
As in the main text, the norm sums absolute differences over pixels and
attribute components, with no additional channel-specific weights.
Only the parameters $\theta$ of the completion Gaussians are optimized;
the retained Gaussians remain fixed.
The targets act as supervision, rather than as additional renderer inputs;
gradients pass through the differentiable rasterizer to the completion
Gaussians.

\paragraph{Geometric meaning of the soft constraints.}
The two penalties in Eq.~(\ref{eq:completion-geometric-losses}) constrain
the normal and tangential displacement of the completion Gaussian centers.
To see their roles, let
$\delta\boldsymbol{\mu}_i=\boldsymbol{\mu}_i-\boldsymbol{\mu}_i^0$.
Because the initial center lies on the plane and the plane frame is
orthonormal, its displacement decomposes as
\begin{equation}
  \begin{aligned}
    \delta\boldsymbol{\mu}_i
      &=\delta u_i\mathbf{e}_u+\delta v_i\mathbf{e}_v
        +\delta h_i\mathbf{n},\\
    (\delta u_i,\delta v_i)^{\top}
      &=T(\boldsymbol{\mu}_i)-\mathbf{u}_i^0,
    \qquad
    \delta h_i=\mathbf{n}^{\top}\boldsymbol{\mu}_i+d.
  \end{aligned}
  \label{eq:completion-displacement-decomposition}
\end{equation}
Thus, $\mathcal{L}_{\mathrm{surf}}$ penalizes normal displacement
$|\delta h_i|$, while $\mathcal{L}_{\mathrm{anchor}}$ penalizes tangential
displacement $|\delta u_i|+|\delta v_i|$.
The anchor is measured in plane coordinates with world units, rather than
texel indices, so its geometric meaning does not depend on an index scaling.
The fitted plane and initial texture coordinates serve as fixed
geometric references during completion optimization.

The surface penalty is a soft geometric constraint.
It allows local departures from the recovered plane when supported by the
reprojected intrinsic supervision, while penalizing excessive deviation.
The anchor likewise permits local tangential adjustment while discouraging
drift from the texture location that supplied the initial appearance.
Neither loss prescribes the exact optimized center.
Both terms constrain Gaussian centers. In particular, the surface term
measures center-to-plane distance; it does not directly penalize Gaussian
orientation.

\paragraph{Completion optimization.}
We first build the completed textures and their per-view targets, then
initialize $\mathcal{G}_h$ with
Eqs.~\eqref{eq:completion-texture-samples}--\eqref{eq:completion-attribute-initialization}.
During optimization, we rasterize the current intrinsic attributes, evaluate
the masked loss in Eq.~(\ref{eq:intrinsic-completion-loss}), and evaluate the
two geometry terms on the completion Gaussians.
Their weighted sum is exactly the objective in
Eq.~(\ref{eq:completion-optimization-objective}).
The retained Gaussians, completed textures, reprojected targets, and
supervision masks remain fixed; only the new Gaussians are updated to fit
these targets under the geometric penalties.
The weights $\lambda_{\mathrm{surf}}$ and $\lambda_{\mathrm{anchor}}$
control the two soft geometry penalties.

Both geometric terms act as soft penalties, allowing the completion
Gaussians to adjust under intrinsic supervision while discouraging
excessive changes from their geometric references.

\paragraph{Reflection rendering after the edit.}
We recompute incident radiance $L_{i,c}^{\theta}$ by tracing rays through the
current scene $\mathcal{G}_{\mathrm{keep}}\cup\mathcal{G}_h$, where
$\mathcal{G}_h$ denotes the completion Gaussians.
Using its current shading normals, Fresnel parameters, and roughness gives
\begin{equation}
  \begin{aligned}
    G_c^{\theta}
      &=\mathcal{H}\!\left(\mathcal{F}_c^{\theta}\odot L_{i,c}^{\theta},
                          R_c^{\theta}\right),\\
    C_c^{\theta}&=D_c^{\theta}+G_c^{\theta}.
  \end{aligned}
  \label{eq:edited-reflection-rendering}
\end{equation}
In particular, the incident radiance is not a fixed map cached from the
unedited scene.
Reflection changes can extend outside $M_T$, because a reflected ray from
outside the footprint may previously have intersected a removed object.
The footprint mask controls intrinsic completion supervision; it does not
restrict where the renderer can update the reflected appearance.

\subsection{Evaluation Protocol and Metric Implementation}
\label{app:evaluation-protocol}

\paragraph{Inputs and evaluation masks.}
We report PSNR, SSIM, LPIPS, and FID, together with their masked counterparts. Syn-Reflective and Syn-Diffuse are evaluated at $800\times800$ and $1600\times1200$ resolution, respectively (width $\times$ height). Predictions, ground truth, and masks correspond to the same views and dimensions. Images are loaded as RGB tensors in $[0,1]$ without automatic resizing. Each evaluation mask combines all object labels selected for removal. On Syn-Reflective, it additionally includes the removed objects' annotated reflections, rather than the entire reflective surface. All methods use identical masks.

\begin{figure}[htbp]
  \centering
  \includegraphics[width=\linewidth]{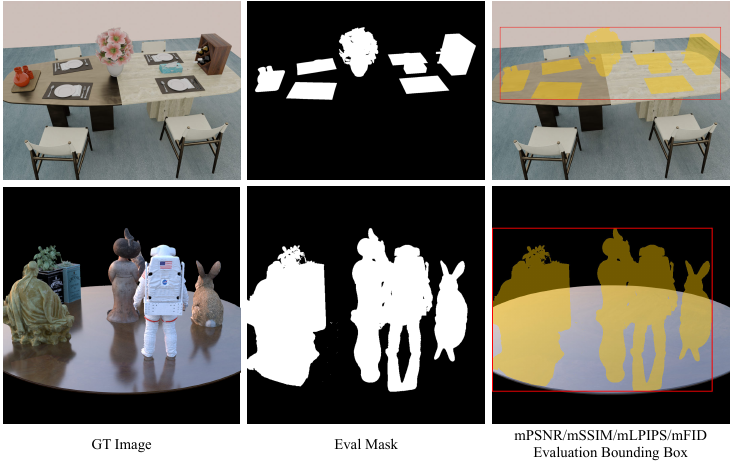}
  \caption{Evaluation regions on Syn-Diffuse (top) and Syn-Reflective (bottom). Columns show the scene image, evaluation mask, and mask overlay with its bounding rectangle (red). All four masked metrics score every pixel inside the rectangle.}
  \label{fig:metric-evaluation-regions}
\end{figure}

\paragraph{Masked evaluation.}
All masked metrics use the smallest axis-aligned bounding rectangle enclosing the nonzero mask pixels, including its boundary. Disconnected regions jointly define one crop. Metrics include all pixels within this rectangle, including unmasked background; no pixelwise mask weighting, zeroing, or ground-truth replacement is applied. Views with empty masks or crops smaller than 32 pixels in either dimension are excluded from masked metrics but retained for full-image evaluation. Fig.~\ref{fig:metric-evaluation-regions} illustrates the evaluation masks and their bounding rectangles for diffuse and reflective scenes.

\paragraph{Metric implementations.}
PSNR uses RGB mean squared error. SSIM uses an $11\times11$ Gaussian window with standard deviation 1.5, five-pixel zero padding, and constants 0.0001 and 0.0009, averaging over channels and spatial locations. LPIPS uses ImageNet-pretrained AlexNet with v0.1 calibration weights and five feature layers. Following the implementation, $[0,1]$ inputs are passed directly without conversion to $[-1,1]$; internal normalization uses mean $(-0.030,-0.088,-0.188)$ and standard deviation $(0.458,0.448,0.450)$. FID uses pytorch-fid 0.3.0 with the FID-specific Inception-v3 and 2048-dimensional pooled features. Its internal preprocessing bilinearly resizes images to $299\times299$ with \texttt{align\_corners=False} and maps values to $[-1,1]$. Masked FID crops are converted to PIL images, saved as PNGs, and reloaded for evaluation. Both FID variants use batch size 1 and eight loading workers. All methods share fixed pretrained metric weights without fine-tuning; FID uses the TensorFlow-converted FID weights rather than standard torchvision classification weights.

\paragraph{Postprocessing and aggregation.}
Saved predictions are evaluated without color or exposure correction, histogram matching, denoising, sharpening, smoothing, registration, or mask morphology. The optional 100-pixel border crop is disabled; metric-internal preprocessing remains enabled. PSNR, SSIM, and LPIPS and their masked variants are averaged over valid views within each scene, then equally over scenes; PSNR is converted to decibels before averaging. FID and masked FID are computed once per dataset by pooling full images or valid crops across scenes, rather than averaging individual image or scene scores.

\onecolumn
\section{Supplementary Results}
\label{app:supplementary-results}

\subsection{Comparisons on Additional Scenes}
\label{app:additional-scenes}

\begin{figure}[H]
  \centering
  \includegraphics[width=\linewidth]{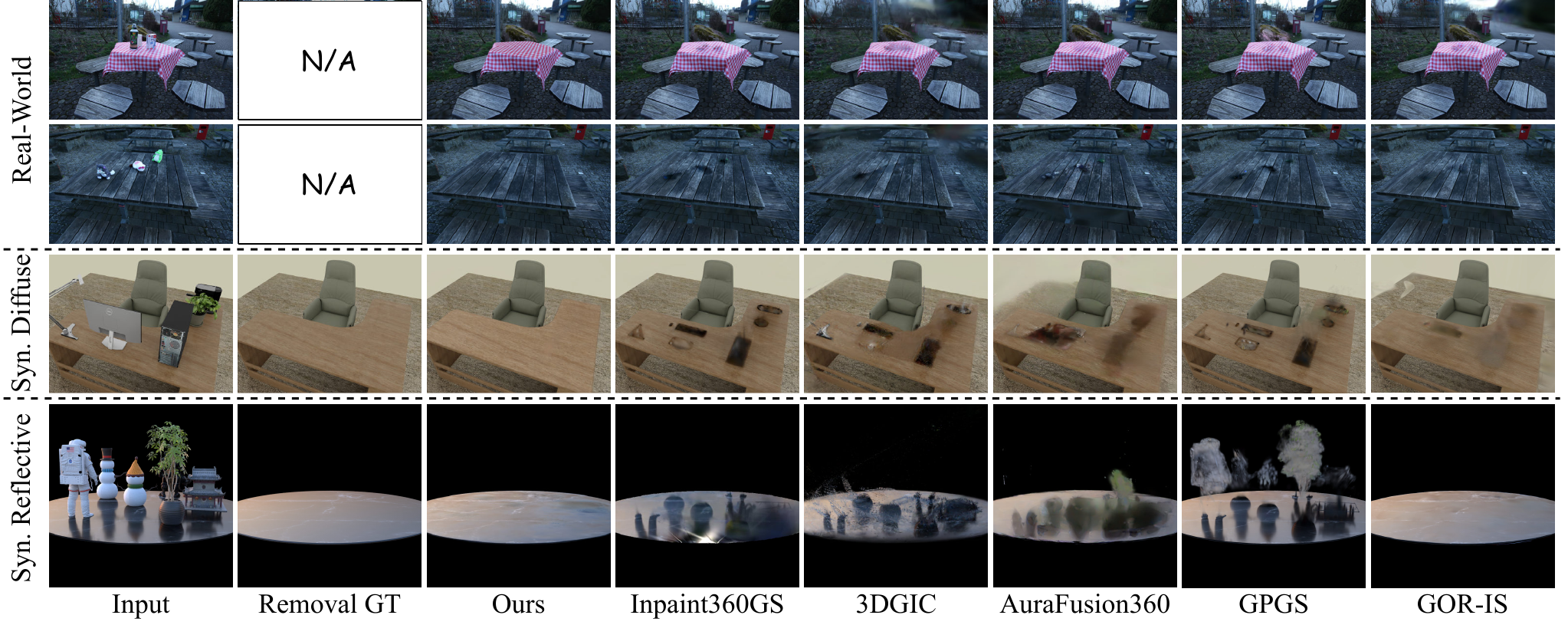}
  \caption{Visual comparison of multi-object removal.}
  \label{fig:multi-appendix}
\end{figure}

Fig.~\ref{fig:multi-appendix} provides additional qualitative comparisons on real-world and synthetic scenes with diffuse and reflective surfaces. Compared with the baselines, 3DTexMOR produces cleaner completions with fewer artifacts and less blurring, preserving the texture and structure of the surrounding background.

\subsection{Ablation on the Number of Removed Objects}
\label{app:removal-count-ablation}

\begin{figure}[H]
  \centering
  \includegraphics[width=\linewidth]{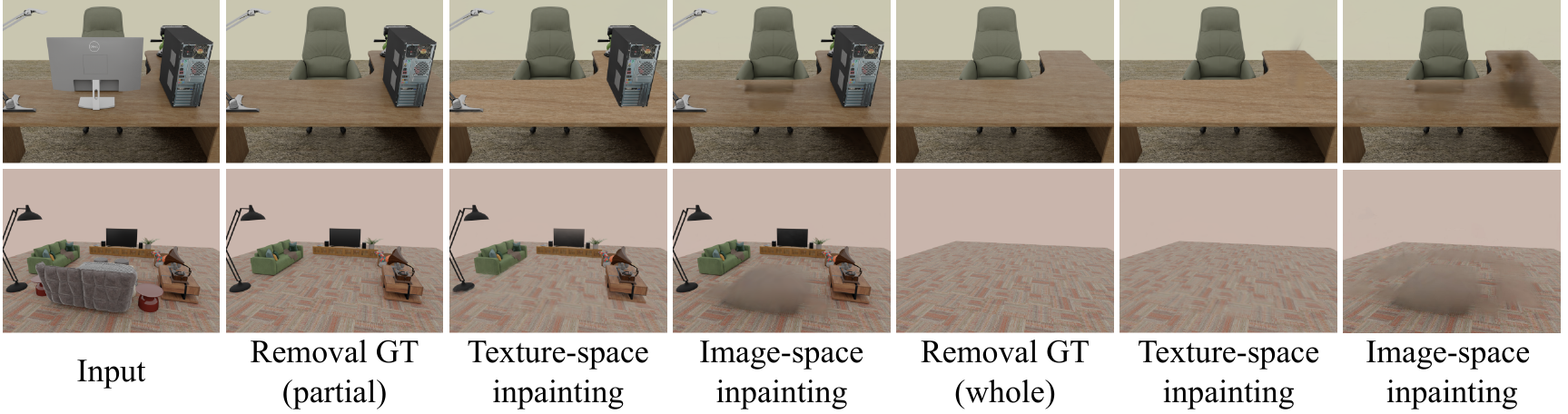}
  \caption{Texture-space and image-space inpainting for single-object and multi-object removal.}
  \label{fig:single-vs-multi}
\end{figure}

We provide supplementary comparisons for the texture-space inpainting
ablation in the main text.
Fig.~\ref{fig:single-vs-multi} compares screen-space and texture-space
inpainting for single-object and multi-object removal, complementing
the texture-space ablation in Fig.~\ref{fig:texture-vs-screen} and
Table~\ref{tab:texture-vs-screen} in the main text.
Image-space inpainting introduces gray artifacts into the inpainted regions, with more artifacts appearing as the number of removed objects increases. Texture-space inpainting avoids these gray artifacts in the illustrated examples, producing cleaner completions for both single-object and multi-object removal.

\clearpage
\subsection{Limitations: Texture-Space Inpainting on Multiple and Complex Surfaces}
\label{app:multi-complex-surface-limitations}
To examine performance beyond approximately planar backgrounds, we select two scenes from the SPIn-NeRF dataset~\citep{mirzaei2023spinnerf}. In scene2, the task is to remove a box in front of uneven ground and a tree trunk, representing a complex, nonplanar background. In scene4, the task is to remove a box from a staircase, whose background consists of multiple planar surfaces.

\paragraph{Separate texture spaces for different background surfaces.}
We expect completion quality to degrade as the background departs from the planar assumption. For backgrounds composed of multiple planes, a possible extension is to assign distinct identity features to background Gaussians using SAM~2~\citep{ravi2024sam2} masks, estimate a separate texture space for each surface, and independently inpaint and reproject the completed textures. However, when the background cannot be decomposed into planar surfaces, as in scene2, the estimated texture surfaces can deviate from the underlying geometry, creating gaps during reprojection.

\begin{figure}[H]
  \centering
  \includegraphics[width=\linewidth]{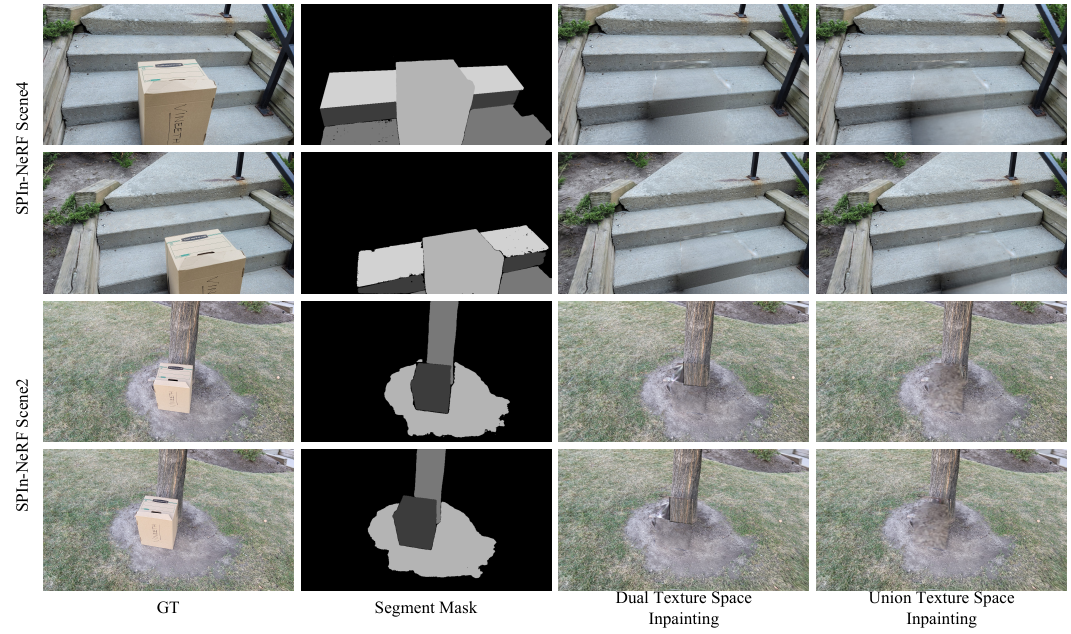}
  \caption{Texture-space inpainting on complex and multi-plane backgrounds from SPIn-NeRF. Dual constructs separate texture spaces for the background categories shown in Segment Mask; Union constructs one shared texture space. Scene4 contains a staircase, while scene2 contains uneven ground and a tree trunk.}
  \label{fig:dual-union-texture-space}
\end{figure}

Fig.~\ref{fig:dual-union-texture-space} compares this extension with a single shared texture space. The \emph{Segment Mask} column shows the background categories learned by the feature-augmented Gaussian field using automatically generated SAM~2 masks. \emph{Dual Texture Space Inpainting} constructs separate texture spaces for the two background categories, whereas \emph{Union Texture Space Inpainting} fits one texture space to their union. Both variants recover parts of the occluded background, suggesting that texture-space inpainting retains some completion capability in these more complex scenes. In scene4, Union produces a conspicuous dark band on the left side of the steps, while Dual gives a cleaner completion. In scene2, the nonplanar trunk leads to a visible gap in the Dual result, whereas Union produces a blurrier completion.

\begin{figure}[H]
  \centering
  \includegraphics[width=0.90\linewidth]{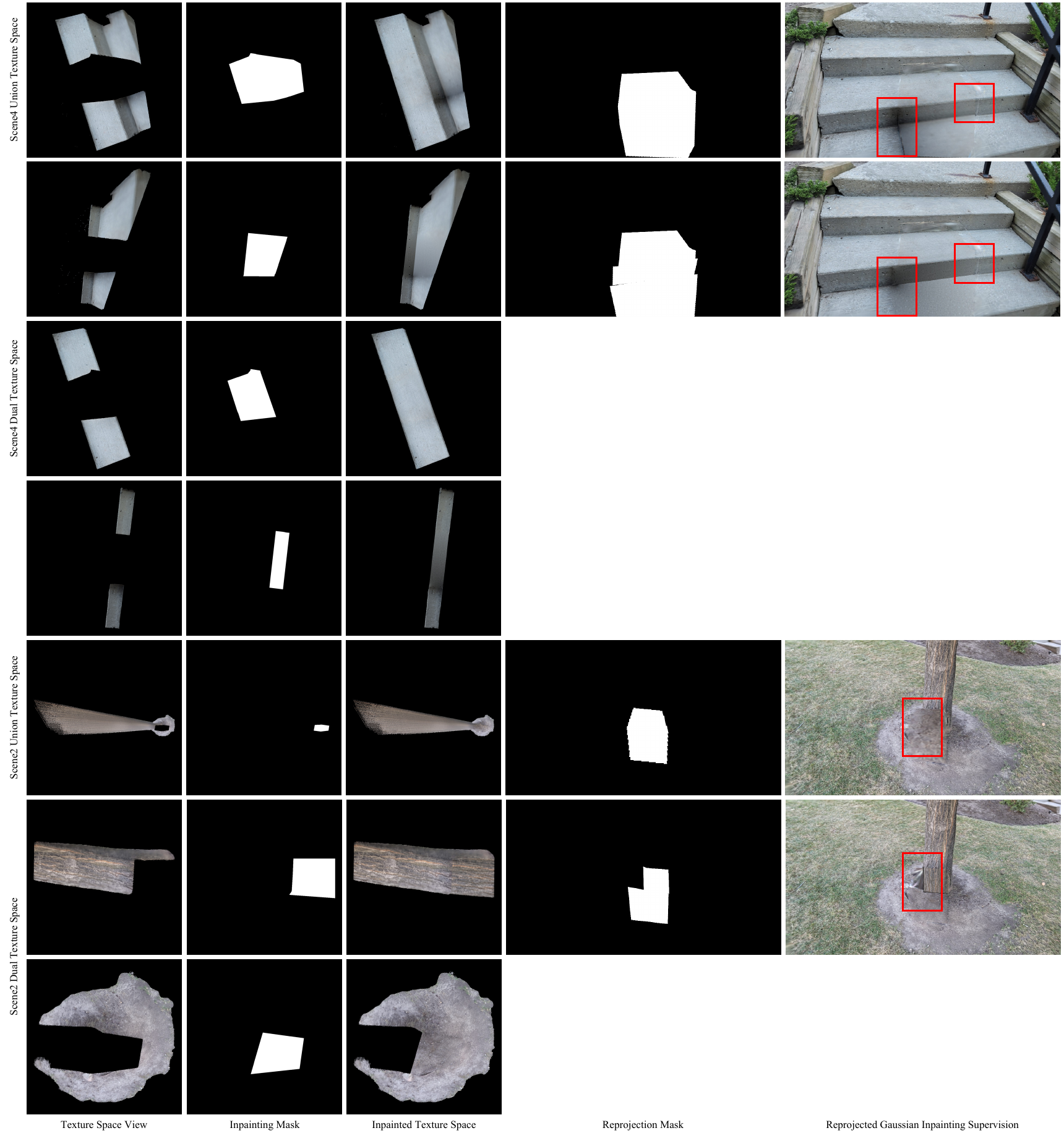}
  \caption{Texture-space completion and reprojected supervision for the Dual and Union variants on scene4 and scene2. Columns show the texture-space view, inpainting mask, completed texture, camera-view reprojection mask, and reprojected inpainting supervision. Red boxes highlight residual artifacts and reprojection mismatches.}
  \label{fig:texture-space-reprojection}
\end{figure}

\paragraph{Texture completion and reprojection failure modes.}
Fig.~\ref{fig:texture-space-reprojection} visualizes the individual or union texture spaces, their inpainting masks, the completed textures, and the reprojection masks and resulting supervision in camera views. The latter is obtained by reprojecting completed textures to supervise the Gaussians in the removal regions across views. The red boxes highlight residual artifacts and mismatches.

In scene4, white patches near the removed object's silhouette remain when residual floaters lie outside the coverage of the reprojected supervision. Since our method first completes a view-independent texture space and then reprojects it, subsequent Gaussian completion cannot remove these floaters through this supervision alone. In scene2, Union maps the uneven ground and trunk into one texture space, which fails to provide appropriate trunk texture: the trunk region is instead filled with ground-like appearance. Separating the surfaces preserves their distinct appearance but introduces gaps where the trunk's nonplanar geometry is not matched by the estimated texture surface. These results expose limitations in both surface parameterization and reprojection coverage on complex backgrounds.

{\small
\begin{thebibliography}{31}
\providecommand{\natexlab}[1]{#1}
\providecommand{\url}[1]{\texttt{#1}}
\expandafter\ifx\csname urlstyle\endcsname\relax
  \providecommand{\doi}[1]{doi: #1}\else
  \providecommand{\doi}{doi: \begingroup \urlstyle{rm}\Url}\fi

\bibitem[Barron et~al.(2022)Barron, Mildenhall, Verbin, Srinivasan, and
  Hedman]{barron2022mipnerf360}
Jonathan~T Barron, Ben Mildenhall, Dor Verbin, Pratul~P Srinivasan, and Peter
  Hedman.
\newblock Mip-nerf 360: Unbounded anti-aliased neural radiance fields.
\newblock In \emph{2022 IEEE/CVF Conference on Computer Vision and Pattern
  Recognition (CVPR)}, pages 5460--5469. IEEE, 2022.

\bibitem[Cen et~al.(2023)Cen, Zhou, Fang, Yang, Shen, Xie, Jiang, Zhang, and
  Tian]{cen2023sa3d}
Jiazhong Cen, Zanwei Zhou, Jiemin Fang, Chen Yang, Wei Shen, Lingxi Xie,
  Dongsheng Jiang, Xiaopeng Zhang, and Qi Tian.
\newblock Segment anything in 3d with nerfs.
\newblock \emph{Advances in Neural Information Processing Systems},
  36:\penalty0 25971--25990, 2023.

\bibitem[Cen et~al.(2025)Cen, Fang, Yang, Xie, Zhang, Shen, and
  Tian]{cen2025saga}
Jiazhong Cen, Jiemin Fang, Chen Yang, Lingxi Xie, Xiaopeng Zhang, Wei Shen, and
  Qi Tian.
\newblock Segment any 3d gaussians.
\newblock In \emph{Proceedings of the AAAI conference on artificial
  intelligence}, pages 1971--1979, 2025.

\bibitem[Chen et~al.(2024)Chen, Chen, Zhang, Wang, Yang, Wang, Cai, Yang, Liu,
  and Lin]{chen2024gaussianeditor}
Yiwen Chen, Zilong Chen, Chi Zhang, Feng Wang, Xiaofeng Yang, Yikai Wang,
  Zhongang Cai, Lei Yang, Huaping Liu, and Guosheng Lin.
\newblock Gaussianeditor: Swift and controllable 3d editing with gaussian
  splatting.
\newblock In \emph{2024 IEEE/CVF Conference on Computer Vision and Pattern
  Recognition (CVPR)}, pages 21476--21485. IEEE, 2024.

\bibitem[Dahaghin et~al.(2026)Dahaghin, Padalkar, Toso, Del~Bue, and
  Murino]{dahaghin2025splatfill}
Mahtab Dahaghin, Milind~G Padalkar, Matteo Toso, Alessio Del~Bue, and Vittorio
  Murino.
\newblock Splatfill: 3d scene inpainting via depth-guided gaussian splatting.
\newblock In \emph{International Conference on Pattern Recognition}, pages
  19--33. Springer, 2026.

\bibitem[Heusel et~al.(2017)Heusel, Ramsauer, Unterthiner, Nessler, and
  Hochreiter]{heusel2017fid}
Martin Heusel, Hubert Ramsauer, Thomas Unterthiner, Bernhard Nessler, and Sepp
  Hochreiter.
\newblock Gans trained by a two time-scale update rule converge to a local nash
  equilibrium.
\newblock \emph{Advances in neural information processing systems}, 30, 2017.

\bibitem[Huang et~al.(2025)Huang, Chou, and Wang]{huang2025_3dgic}
Sheng-Yu Huang, Zi-Ting Chou, and Yu-Chiang~Frank Wang.
\newblock 3d gaussian inpainting with depth-guided cross-view consistency.
\newblock In \emph{2025 IEEE/CVF Conference on Computer Vision and Pattern
  Recognition (CVPR)}, pages 26704--26713. IEEE, 2025.

\bibitem[Kerbl et~al.(2023)Kerbl, Kopanas, Leimk{\"u}hler, and
  Drettakis]{kerbl2023gaussian}
Bernhard Kerbl, Georgios Kopanas, Thomas Leimk{\"u}hler, and George Drettakis.
\newblock 3d gaussian splatting for real-time radiance field rendering.
\newblock \emph{ACM Trans. Graph.}, 42\penalty0 (4):\penalty0 139:1--139:14,
  2023.

\bibitem[Kerr et~al.(2023)Kerr, Kim, Goldberg, Kanazawa, and
  Tancik]{kerr2023lerf}
Justin Kerr, Chung~Min Kim, Ken Goldberg, Angjoo Kanazawa, and Matthew Tancik.
\newblock Lerf: Language embedded radiance fields.
\newblock In \emph{2023 IEEE/CVF International Conference on Computer Vision
  (ICCV)}, pages 19672--19682. IEEE, 2023.

\bibitem[Kirillov et~al.(2023)Kirillov, Mintun, Ravi, Mao, Rolland, Gustafson,
  Xiao, Whitehead, Berg, Lo, et~al.]{kirillov2023segment}
Alexander Kirillov, Eric Mintun, Nikhila Ravi, Hanzi Mao, Chloe Rolland, Laura
  Gustafson, Tete Xiao, Spencer Whitehead, Alexander~C Berg, Wan-Yen Lo, et~al.
\newblock Segment anything.
\newblock In \emph{2023 IEEE/CVF international conference on computer vision
  (ICCV)}, pages 3992--4003. IEEE, 2023.

\bibitem[Kocour et~al.(2025)Kocour, Benbihi, and Sattler]{kocour2025remove360}
Simona Kocour, Assia Benbihi, and Torsten Sattler.
\newblock Remove360: Benchmarking residuals after object removal in 3d gaussian
  splatting.
\newblock \emph{arXiv preprint arXiv:2508.11431}, 2025.

\bibitem[Lee and Cho(2026)]{lee2026gpgs}
Yongjoon Lee and Donghyeon Cho.
\newblock Gpgs: Consistent 3d object removal via geometry-aware 3d inpainting
  and projected image refinement in 3d gaussian splatting.
\newblock In \emph{Proceedings of the AAAI Conference on Artificial
  Intelligence}, pages 5927--5935, 2026.

\bibitem[Lugmayr et~al.(2022)Lugmayr, Danelljan, Romero, Yu, Timofte, and
  Van~Gool]{lugmayr2022repaint}
Andreas Lugmayr, Martin Danelljan, Andres Romero, Fisher Yu, Radu Timofte, and
  Luc Van~Gool.
\newblock Repaint: Inpainting using denoising diffusion probabilistic models.
\newblock In \emph{2022 IEEE/CVF conference on computer vision and pattern
  recognition (CVPR)}, pages 11451--11461. IEEE, 2022.

\bibitem[Mildenhall et~al.(2021)Mildenhall, Srinivasan, Tancik, Barron,
  Ramamoorthi, and Ng]{mildenhall2020nerf}
Ben Mildenhall, Pratul~P Srinivasan, Matthew Tancik, Jonathan~T Barron, Ravi
  Ramamoorthi, and Ren Ng.
\newblock Nerf: Representing scenes as neural radiance fields for view
  synthesis.
\newblock \emph{Communications of the ACM}, 65\penalty0 (1):\penalty0 99--106,
  2021.

\bibitem[Mirzaei et~al.(2023)Mirzaei, Aumentado-Armstrong, Derpanis, Kelly,
  Brubaker, Gilitschenski, and Levinshtein]{mirzaei2023spinnerf}
Ashkan Mirzaei, Tristan Aumentado-Armstrong, Konstantinos~G Derpanis, Jonathan
  Kelly, Marcus~A Brubaker, Igor Gilitschenski, and Alex Levinshtein.
\newblock Spin-nerf: Multiview segmentation and perceptual inpainting with
  neural radiance fields.
\newblock In \emph{2023 IEEE/CVF Conference on Computer Vision and Pattern
  Recognition (CVPR)}, pages 20669--20679. IEEE, 2023.

\bibitem[Qin et~al.(2024)Qin, Li, Zhou, Wang, and Pfister]{qin2024langsplat}
Minghan Qin, Wanhua Li, Jiawei Zhou, Haoqian Wang, and Hanspeter Pfister.
\newblock Langsplat: 3d language gaussian splatting.
\newblock In \emph{2024 IEEE/CVF Conference on Computer Vision and Pattern
  Recognition (CVPR)}, pages 20051--20060. IEEE, 2024.

\bibitem[Ravi et~al.(2025)Ravi, Gabeur, Hu, Hu, Ryali, Ma, Khedr, R{\"a}dle,
  Rolland, Gustafson, et~al.]{ravi2024sam2}
Nikhila Ravi, Valentin Gabeur, Yuan-Ting Hu, Ronghang Hu, Chaitanya Ryali,
  Tengyu Ma, Haitham Khedr, Roman R{\"a}dle, Chloe Rolland, Laura Gustafson,
  et~al.
\newblock Sam 2: Segment anything in images and videos.
\newblock In \emph{International Conference on Learning Representations}, pages
  28085--28128, 2025.

\bibitem[Seo et~al.(2026)Seo, Yoo, and Kim]{seo2025repaintgs}
Ji~Hyun Seo, Byounhyun Yoo, and Gerard~Jounghyun Kim.
\newblock Reference-guided gaussian splatting for realistic and view-consistent
  3d scene inpainting.
\newblock \emph{Computational Visual Media}, 2026.

\bibitem[Suvorov et~al.(2022)Suvorov, Logacheva, Mashikhin, Remizova, Ashukha,
  Silvestrov, Kong, Goka, Park, and Lempitsky]{suvorov2022lama}
Roman Suvorov, Elizaveta Logacheva, Anton Mashikhin, Anastasia Remizova,
  Arsenii Ashukha, Aleksei Silvestrov, Naejin Kong, Harshith Goka, Kiwoong
  Park, and Victor Lempitsky.
\newblock Resolution-robust large mask inpainting with fourier convolutions.
\newblock In \emph{2022 IEEE/CVF Winter Conference on Applications of Computer
  Vision (WACV)}, pages 3172--3182. IEEE, 2022.

\bibitem[Wang et~al.(2026)Wang, Zhang, Millerdurai, Westermann, Stricker, and
  Pagani]{wang2026inpaint360gs}
Shaoxiang Wang, Shihong Zhang, Christen Millerdurai, R{\"u}diger Westermann,
  Didier Stricker, and Alain Pagani.
\newblock Inpaint360gs: Efficient object-aware 3d inpainting via gaussian
  splatting for 360° scenes.
\newblock In \emph{2026 IEEE/CVF Winter Conference on Applications of Computer
  Vision (WACV)}, pages 117--127. IEEE, 2026.

\bibitem[Wang et~al.(2024)Wang, Wu, Zhang, and Xu]{wang2024gscream}
Yuxin Wang, Qianyi Wu, Guofeng Zhang, and Dan Xu.
\newblock Learning 3d geometry and feature consistent gaussian splatting for
  object removal.
\newblock In \emph{European conference on computer vision}, pages 1--17.
  Springer, 2024.

\bibitem[Wang et~al.(2004)Wang, Bovik, Sheikh, and Simoncelli]{wang2004ssim}
Zhou Wang, Alan~C Bovik, Hamid~R Sheikh, and Eero~P Simoncelli.
\newblock Image quality assessment: from error visibility to structural
  similarity.
\newblock \emph{IEEE transactions on image processing}, 13\penalty0
  (4):\penalty0 600--612, 2004.

\bibitem[Weber et~al.(2024)Weber, Holynski, Jampani, Saxena, Snavely, Kar, and
  Kanazawa]{weber2024nerfiller}
Ethan Weber, Aleksander Holynski, Varun Jampani, Saurabh Saxena, Noah Snavely,
  Abhishek Kar, and Angjoo Kanazawa.
\newblock Nerfiller: Completing scenes via generative 3d inpainting.
\newblock In \emph{2024 IEEE/CVF Conference on Computer Vision and Pattern
  Recognition (CVPR)}, pages 20731--20741. IEEE, 2024.

\bibitem[Weder et~al.(2023)Weder, Garcia-Hernando, Monszpart, Pollefeys,
  Brostow, Firman, and Vicente]{weder2023removing}
Silvan Weder, Guillermo Garcia-Hernando, Aron Monszpart, Marc Pollefeys,
  Gabriel~J Brostow, Michael Firman, and Sara Vicente.
\newblock Removing objects from neural radiance fields.
\newblock In \emph{Proceedings of the IEEE/CVF Conference on Computer Vision
  and Pattern Recognition}, pages 16528--16538, 2023.

\bibitem[Wu et~al.(2025)Wu, Chen, Chen, Lee, Ke, Mu, Huang, Lin, Chen, Lin,
  et~al.]{wu2025aurafusion360}
Chung-Ho Wu, Yang-Jung Chen, Ying-Huan Chen, Jie-Ying Lee, Bo-Hsu Ke,
  Chun-Wei~Tuan Mu, Yi-Chuan Huang, Chin-Yang Lin, Min-Hung Chen, Yen-Yu Lin,
  et~al.
\newblock Aurafusion360: Augmented unseen region alignment for reference-based
  360 unbounded scene inpainting.
\newblock In \emph{2025 IEEE/CVF Conference on Computer Vision and Pattern
  Recognition (CVPR)}, pages 16366--16376. IEEE, 2025.

\bibitem[Ye et~al.(2024)Ye, Danelljan, Yu, and Ke]{ye2024gaussiangrouping}
Mingqiao Ye, Martin Danelljan, Fisher Yu, and Lei Ke.
\newblock Gaussian grouping: Segment and edit anything in 3d scenes.
\newblock In \emph{European conference on computer vision}, pages 162--179.
  Springer, 2024.

\bibitem[You et~al.(2025)You, Lin, Lyu, Zhang, and Yang]{you2025instainpaint}
Junqi You, Chieh Lin, Weijie Lyu, Zhengbo Zhang, and Ming-Hsuan Yang.
\newblock Instainpaint: Instant 3d-scene inpainting with masked large
  reconstruction model.
\newblock \emph{Advances in Neural Information Processing Systems},
  38:\penalty0 153574--153594, 2025.

\bibitem[Zhang et~al.(2018)Zhang, Isola, Efros, Shechtman, and
  Wang]{zhang2018lpips}
Richard Zhang, Phillip Isola, Alexei~A Efros, Eli Shechtman, and Oliver Wang.
\newblock The unreasonable effectiveness of deep features as a perceptual
  metric.
\newblock In \emph{2018 IEEE/CVF conference on computer vision and pattern
  recognition}, pages 586--595. IEEE, 2018.

\bibitem[Zhao et~al.(2026)Zhao, Gao, Yang, Xie, and Wang]{zhao2026goris}
Yonghao Zhao, Yupeng Gao, Jian Yang, Jin Xie, and Beibei Wang.
\newblock Gor-is: 3d gaussian object removal in the intrinsic space.
\newblock In \emph{Proceedings of the IEEE/CVF Conference on Computer Vision
  and Pattern Recognition}, pages 40896--40906, 2026.

\bibitem[Zheng et~al.(2026)Zheng, Li, Xiang, Peng, and Wang]{zheng2026gaucomp}
Wenxing Zheng, Yuqi Li, Jinghui Xiang, Xiaohao Peng, and Chong Wang.
\newblock Gaucomp: 3d gaussian completion for associated shadow and object
  removal.
\newblock \emph{Computer Graphics Forum}, page e70309, 2026.

\bibitem[Zhou et~al.(2024)Zhou, Chang, Jiang, Fan, Zhu, Xu, Chari, You, Wang,
  and Kadambi]{zhou2024feature3dgs}
Shijie Zhou, Haoran Chang, Sicheng Jiang, Zhiwen Fan, Zehao Zhu, Dejia Xu,
  Pradyumna Chari, Suya You, Zhangyang Wang, and Achuta Kadambi.
\newblock Feature 3dgs: Supercharging 3d gaussian splatting to enable distilled
  feature fields.
\newblock In \emph{2024 IEEE/CVF Conference on Computer Vision and Pattern
  Recognition (CVPR)}, pages 21676--21685. IEEE, 2024.

\end{thebibliography}
}
\end{document}